\documentclass[11pt]{article}

\usepackage[a4paper,margin=1in]{geometry}
\usepackage[T1]{fontenc}
\usepackage[utf8]{inputenc}
\usepackage[numbers,sort&compress]{natbib}
\usepackage{graphicx}
\usepackage{amsmath}
\usepackage{algorithm}
\usepackage{algorithmic}
\usepackage{listings}
\usepackage{booktabs}
\usepackage{array}
\usepackage[hidelinks]{hyperref}

\newcommand{\tblwidth}{\textwidth}

\title{MAIL: Memory-driven, Adaptive, Incremental, and Literature-grounded Framework for Hypothesis Generation in Chemistry}

\author{
Mahdi Babaei$^{1}$, Xueshen Li$^{1}$, Yutao Kuang$^{2}$,\\
Jolene P. Reid$^{2,*}$, and Yu Gan$^{1,3,4,*}$\\[0.75em]
\small $^{1}$Biomedical Engineering, Stevens Institute of Technology, Hoboken, NJ 07030, USA\\
\small $^{2}$Department of Chemistry, University of British Columbia, Vancouver, BC V6T 1Z1, Canada\\
\small $^{3}$Department of Bioengineering, University of Maryland, College Park, MD 20742, USA\\
\small $^{4}$Artificial Intelligence Interdisciplinary Institute at Maryland,\\
\small University of Maryland, College Park, MD 20742, USA\\[0.5em]
\small $^{*}$Corresponding authors: jreid@chem.ubc.ca; yugan@umd.edu
}

\date{}

\begin{document}
\maketitle

\begin{abstract}
The ever-expanding volume of the chemical literature offers unprecedented opportunities to generate novel and impactful hypotheses. However, the bottleneck lies in efficiently navigating this vast knowledge base to formulate high-quality, experimentally meaningful insights.
 While Large Language Models (LLMs) show promise for this task, existing methods often rely on static inspiration corpora, predefined heuristics, or laborious human-in-the-loop pipelines and decision-support frameworks that limit scalability and novelty. In this work, we propose an automated approach, a Memory-augmented, Adaptive, Incremental, and Literature-grounded (MAIL) framework for hypothesis generation in chemistry. 
Our MAIL method formulates hypothesis generation as a temporally grounded, memory-driven reasoning process, where hypotheses emerge from an evolving conceptual path that continuously accumulates and reinterprets prior knowledge. We evaluated the MAIL framework on a public TOMATO-Chem dataset and a newly curated and disseminated
high-novelty nature/science challenge (HN-NS) dataset. Across both datasets, MAIL generates structurally coherent and mechanistically plausible hypotheses, achieves the highest MIOS and MPOS by more effectively recovering the central ideas and methodological elements of the historical target hypotheses, and obtains the highest overall expert-evaluation scores for scientific quality. These results demonstrate the potential of LLMs to autonomously explore chemical domains and generate hypotheses that are both innovative and chemically plausible. 
\end{abstract}


\vspace{1em}

\section{Introduction}

The ever-expanding volume of chemical literature prese\-nts an unprecedented opportunity for scientific discovery \citep{white2019publications}. While recent advancements have been achieved in retrieval \citep{noh2024retrieval}, unlocking its full potential in chemical discovery demands the capability to trace how ideas evolve, reason through complex chemical systems, and finally generate novel, actionable hypotheses for future advancement \citep{OTYEPKA2025102981, zhang2025exploring}. Large Language Models (LLMs) have emerged as transformative tools for complex reasoning across scientific disciplines \citep{zhao2024survey}. A compelling application is automated hypothesis generation, where systems like the Zero-shot Proposer \citep{qi2023large} demonstrate how LLMs can synthesize insights from vast scientific corpora to propose novel research directions. 

Natural science hypotheses must adhere to domain-specific constraints, demanding mechanistic precision, structured reasoning, and practical feasibility. This raises fundamental questions about how language models can operate within the rigor of chemical inquiry, ensuring hypotheses align with domain knowledge and remain practically meaningful without real-time experimental feedback \citep{kumar-etal-2025-large, yang2024moose}.

In chemistry, these limitations are particularly important because generated hypotheses must remain consistent with chemical principles while also specifying experimentally meaningful mechanisms and outcomes \citep{mirza2025framework}. Previous research has begun to address different aspects of this problem. Temporal knowledge-graph methods model the evolution of scientific entities and relationships over time \citep{LU20111150,xiong2024improvingscientifichypothesisgeneration}. MOOSE-Chem \citep{yang2024moose} uses multi-stage inspiration retrieval and composition to construct chemistry hypotheses, while SciMON \citep{wang2024scimon} and Scideator \citep{radensky2024scideator} recombine concepts or scientific facets to support hypothesis generation. ChemReasoner \citep{sprueill2402chemreasoner} further applies iterative search with quantum-chemical feedback to catalyst design. More recent approaches include goal-driven and constraint-guided agents for materials hypothesis generation \citep{kumbhar-etal-2025-hypothesis} and hierarchical search for incrementally constructing fine-grained chemistry hypotheses \citep{yang2025moosechem2}.

These studies demonstrate the value of temporal, iterative, and compositional reasoning for scientific hypothesis generation. However, they primarily focus on knowledge-graph evolution, composition from a predefined literature corpus, interactive concept recombination, hierarchical detail expansion, or domain-specific optimization. A remaining gap is the integration of query-dependent retrieval from multiple scientific databases across chronological intervals with compressed cross-round memory and feedback-guided natural-language hypothesis refinement. MAIL addresses this specific gap by allowing the literature retrieved in each round to respond to the hypothesis and feedback produced in preceding rounds.  

\begin{figure}
    \centering
    \includegraphics[width=0.99\columnwidth]{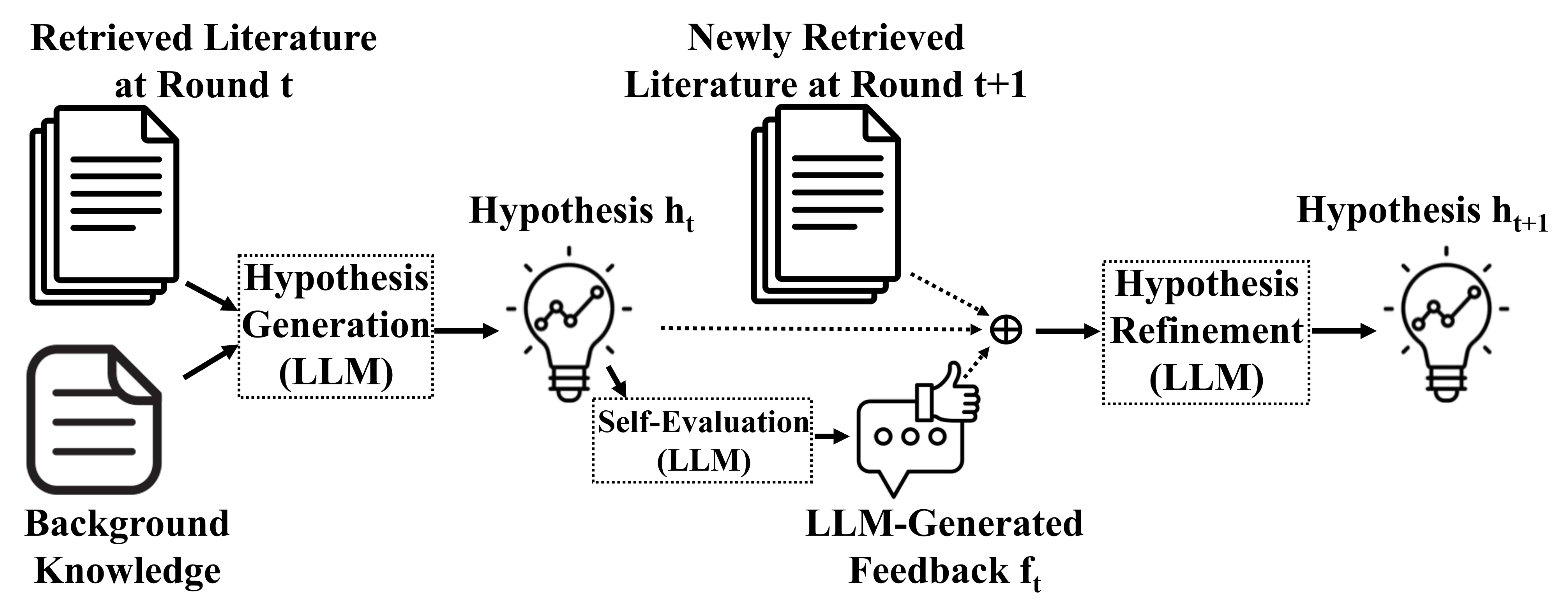}
    \caption{
High-level overview of the proposed autonomous hypothesis generation method. At round $t$, the research background and retrieved literature are used to generate the current hypothesis $h_t$. The LLM then evaluates this hypothesis and produces internal feedback $f_t$. In the following round, the current hypothesis, LLM-generated feedback, and newly retrieved literature $I^{(t+1)}$ are jointly used to generate the refined hypothesis $h_{t+1}$. This iterative process continues without human feedback until the final hypothesis is selected.
}
    \label{fig1}
\end{figure}

In this paper, we design a Memory-augmented, Adaptive, Incremental, and Literature-grounded (MAIL) framework. The framework eliminates the need for manually selected inspiration sources, heuristic decompositions, or handcrafted ranking procedures. As shown in Fig.~\ref{fig1}, we cast hypothesis generation as a temporally grounded, memory-driven reasoning task, where each hypothesis emerges not in isolation but as part of an evolving conceptual path that accumulates and reinterprets prior knowledge. MAIL incrementally constructs hypotheses by retrieving temporally prior literature and applying a chain of prompts (CoP) \citep{wei2022chain} that guide the LLM to synthesize, refine, and evaluate scientific ideas. At each subsequent round, the current hypothesis is retained and incrementally updated using newly retrieved literature and feedback generated internally by the LLM. Thus, the feedback loop shown in Fig.~\ref{fig1} does not involve human evaluation during inference; it represents the model's self-evaluation of the current hypothesis. Empowered by in-context learning \citep{huang2024multimodal, liu2024context}, the model can adjust its generative behavior, encouraging greater novelty, mechanistic depth, and alignment with domain-specific reasoning.

The major contributions are:
\begin{itemize}
    \item We designed an iterative, dynamic retrieval engine that incorporates compressed memory to augment discovery beyond static or manually curated corpora. 
    \item We design a multi-round hypothesis process using CoP and in-context learning that treats hypothesis formation as a temporal reasoning trajectory. We adaptively incorporate new information from internal feedback. 
    \item We curate and release a high-novelty Nature/Science challenge (HN-NS) dataset to rigorously test automated scientific discovery systems. We validate the generalizability and robustness of MAIL on this newly available dataset alongside the public TOMATO-Chem benchmark.
\end{itemize}

\section{Related Work} 
\subsection{Graph-based Scientific Discovery} 
Automated scientific discovery has evolved from text mining and association rule inference \citep{swanson1986fish, spangler2014automated, belford2018stability} to modern graph-based reasoning. Recent approaches utilize temporal link prediction and dynamic knowledge graphs to model entity evolution \citep{LU20111150, xiong2024improvingscientifichypothesisgeneration, alkan2025surveyhypothesisgenerationscientific}, and formulate hypothesis generation directly as a link prediction task over scientific knowledge graphs to identify previously unknown connections \citep{BORREGO2025113280}, or apply analogical reasoning over knowledge graphs to assist LLMs in idea generation \citep{oyelade2025smar+}. However, these methods often rely on predefined structural patterns and lack the reasoning required for complex chemical systems. Another study proposes advancing abductive reasoning in knowledge graphs through logical hypothesis generation, enabling the identification of structured mechanistic explanations \citep{bai-etal-2024-advancing}.

Compositional scientific discovery has also been explored through reinforcement learning (RL) and Generative Flow Networks (GFlowNets). In RL-based molecular generation, a molecule is constructed as a sequence of actions, such as adding atoms or bonds, while a policy is optimized using domain-specific rewards and chemical constraints. For example, the Graph Convolutional Policy Network (GCPN) formulates goal-directed molecular graph generation as a sequential decision process and optimizes molecular properties through policy-gradient learning \citep{you2018gcpn}. GFlowNets similarly construct objects through action trajectories, but learn a stochastic policy that samples diverse terminal objects approximately in proportion to a specified positive reward \citep{bengio2021gflownet,bengio2023gflownet}. This makes GFlowNets particularly useful when multiple structurally diverse and high-reward molecular candidates are desired.

Although these approaches are effective for molecular and graph-based optimization, they assume that the state space, available actions, structural constraints, and reward function can be explicitly defined. MAIL addresses a different level of scientific discovery: it retrieves and synthesizes information from unstructured scientific literature to generate and refine natural-language research hypotheses. Furthermore, the trajectory in RL and GFlowNets represents the sequential construction of a graph or molecular object, whereas the trajectory in MAIL represents successive literature intervals and corresponding hypothesis-refinement rounds. MAIL is therefore better suited to the open-ended, literature-grounded hypothesis-generation problem considered in this work, where an explicit molecular graph and numerical reward function are generally unavailable. These approaches are complementary rather than mutually exclusive: a hypothesis generated by MAIL could subsequently define objectives or constraints for an RL or GFlowNet-based molecular optimization system.

\subsection{LLMs in Hypothesis Generation} 
Recent work has shifted toward using Large Language Models (LLMs) for scientific hypothesis generation \citep{zheng-etal-2025-automation}. MOOSE \citep{yang2023large} introduced inspiration-based hypothesis generation in the social sciences, while MOOSE-Chem \citep{yang2024moose} extended this formulation to chemistry through multi-stage inspiration retrieval, composition, and ranking. SciMON \citep{wang2024scimon} and Scideator \citep{radensky2024scideator} explore hypothesis generation through concept or facet recombination. Other approaches combine literature-derived theoretical relationships with empirical data patterns \citep{liu-etal-2025-literature}, or employ search and scientific feedback for specialized chemical problems, as demonstrated by NOVA \citep{hu2025nova} and ChemReasoner \citep{sprueill2402chemreasoner}. Goal-driven and constraint-guided LLM agents have also been proposed for materials hypothesis generation \citep{kumbhar-etal-2025-hypothesis}. MOOSE-Chem2 \citep{yang2025moosechem2} further formulates fine-grained chemistry hypothesis generation as a hierarchical search process that incrementally adds methodological and experimental details.

These methods establish important compositional, iterative, and search-based foundations for scientific hypothesis generation. MAIL differs in its specific integration of chronological literature partitioning, query-dependent retrieval from multiple external databases, compressed memory of previous rounds, and feedback-guided hypothesis refinement. Thus, MAIL does not replace these approaches but addresses the complementary problem of maintaining temporal literature grounding and conceptual continuity across multiple refinement rounds.

\subsection{Reasoning and Retrieval Strategies} 
General reasoning techniques, including Chain of Thought \citep{wang2024chain}, Tree-of-Thought \citep{yao2023tree}, self-reflection loops \citep{shinn2023reflexion}, and selective prompting strategies paired with chain-of-thought \citep{che2026select}, improve LLM coherence but often lack domain-specific literature grounding. Standard Retrieval-Augmented Generation (RAG) \citep{lewis2021retrievalaugmentedgenerationknowledgeintensivenlp, guu2020realmretrievalaugmentedlanguagemodel} mitigates hallucinations but typically operates on static datasets. Recent integrations of RAG with knowledge graphs \citep{10.1093/bioinformatics/btae353, delile2024graphbasedretrievercaptureslong} enhance retrieval across domains. However, they lack mechanisms for the iterative, literature-grounded hypothesis refinement and memory compression.

\begin{figure}[t]
\centering
\includegraphics[width=1\textwidth]{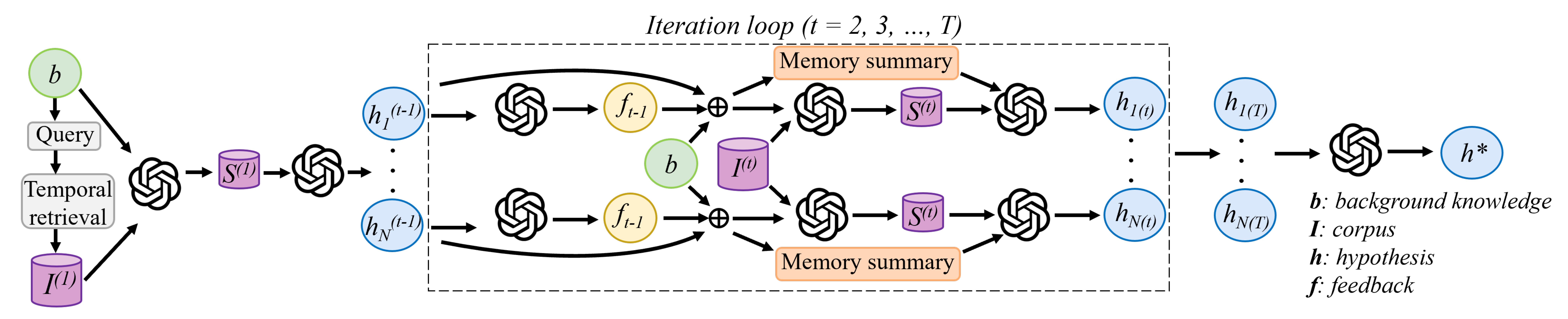}
\caption{
Overview of the MAIL framework. Starting from the fixed background $b$,
the model constructs a query and performs temporally filtered literature
retrieval to obtain $I^{(1)}$. The selected inspirations $S^{(1)}$ are
then used to initialize $N$ candidate hypothesis trajectories. At each
subsequent round $t$, the preceding hypothesis $h_j^{(t-1)}$ is
self-evaluated to produce structured feedback $f_{t-1}$. This feedback,
together with $b$, the dynamically retrieved literature $I^{(t)}$, and
a compressed memory summary of previously selected inspirations, guides
the selection of $S^{(t)}$ and refinement of each candidate into
$h_j^{(t)}$. After $T$ rounds, a target-blind selection prompt chooses
the final hypothesis $h^*$ from the refined candidates. All feedback,
refinement, and selection operations are performed automatically without
human intervention during inference.
}
\label{fig2}
\end{figure}

\section{Methodology}
Our main assumption is that novel hypotheses can be systematically constructed through multi-round interactions between a defined background problem and a diverse set of inspiration sources drawn from temporally preceding literature. Scientific hypothesis generation in chemistry is not a one-shot process. Rather, we model the hypothesis generation as a dynamic reasoning trajectory, one that unfolds over multiple stages, builds on prior knowledge, revisits inspiration with new context, and refines emerging ideas over time. This mimics how chemists often arrive at impactful hypotheses: through a sequence of conceptual refinements shaped by accumulated evidence and evolving theoretical perspectives.

As shown in Fig.~\ref{fig2}, the MAIL decomposes the hypothesis generation into three interdependent components: (1) literature-based inspiration retrieval, (2) incremental refinement with self-evaluation, and (3) prompt adaptation and hypothesis selection. Inspired by a chain of thoughts, these components operate over a sequence of rounds, each building on earlier outputs to generate hypotheses of increasing specificity, novelty, and coherence. Each round incorporates more recent literature, enabling temporally grounded and forward-propagated reasoning. 

Let $b$ denote the fixed background provided for each target instance, consisting of a research question and its accompanying background survey. MAIL does not assume a predefined literature corpus. Instead, at each round $t$, the model constructs a query $q_t$ and executes it through the PubMed, CrossRef, and Semantic Scholar APIs using the publication interval $\Delta_t$ as a date filter. The returned collection of article records is denoted by $I^{(t)}$:
\begin{equation}
q_1=\mathcal{Q}(b), \qquad
q_t=\mathcal{Q}(b,h_{t-1},f_{t-1}),\quad t=2,\ldots,T,
\end{equation}
\begin{equation}
I^{(t)}=\operatorname{Retrieve}(q_t,\Delta_t).
\end{equation}
Each $I^{(t)}$ therefore contains a query-dependent collection of articles retrieved during round $t$; it does not contain sections obtained by segmenting an individual article. We use
\begin{equation}
I=\bigcup_{t=1}^{T}I^{(t)}
\end{equation}
only as shorthand for all literature records retrieved across the complete refinement process, rather than as a predefined input corpus.
Thus, $b$ and $I$ are not the same: $b$ is the fixed task description supplied to the model, whereas $I$ denotes the article records dynamically retrieved using queries generated from that task description and the evolving refinement state.

New inspirations are selected from the dynamically retrieved article set rather than generated directly from $b$. At $t=1$, the background $b$ is used to construct the initial query $q_1$. Executing this query with the date filter $\Delta_1$ produces $I^{(1)}$, from which the model selects the most conceptually or mechanistically useful articles as $S^{(1)}$. For $t>1$, query construction and inspiration selection are additionally conditioned on the previous hypothesis $h_{t-1}$ and its feedback $f_{t-1}$. The selection process is represented as
\begin{equation}
\begin{aligned}
S^{(1)} &= \operatorname{Sel}(I^{(1)}\mid b),\\
S^{(t)} &= \operatorname{Sel}
(I^{(t)}\mid b,h_{t-1},f_{t-1}),
\end{aligned}
\label{eq:inspiration_selection}
\end{equation}
The selected articles are subsequently used to generate or refine the hypothesis. Formally, we define this recursive hypothesis refinement as:

\begin{equation}
\label{eq:1}
h_t = \mathcal{G}(b, h_{t-1}, f_{t-1}, S^{(t)}), \quad t = 2, \ldots, T .
\end{equation}
Where $\mathcal{G}$ is the hypothesis generation function, which is implemented via an LLM. The base case uses no prior hypotheses:

\begin{equation}
\label{eq:2}
h_1 = \mathcal{G}(b, \emptyset, \emptyset, S^{(1)}),
\end{equation}
with inspirations $S^{(1)}$ selected using only the background information $b$. The round index begins at $t=1$; therefore, $I^{(1)}$, $S^{(1)}$, and $h_1$ correspond to the first refinement round, while $h_0=f_0=\emptyset$ represents the absence of a previous hypothesis and feedback.
For clarity, Equations~\ref{eq:1} and~\ref{eq:2} describe one candidate refinement trajectory. More generally, let $h_j^{(t)}$ denote the $j$th candidate hypothesis after round $t$, and let
\begin{equation}
H^{(t)}=\left\{h_j^{(t)}\right\}_{j=1}^{N},
\end{equation}
where $N$ is a configurable candidate-generation budget. The superscript denotes the refinement round, whereas the subscript denotes the candidate trajectory. After the final round, the hypothesis-selection prompt selects the final output from the refined candidate set:
\begin{equation}
h^*=\mathcal{A}\!\left(H^{(T)}\mid b\right),
\end{equation}
where $\mathcal{A}$ denotes the final hypothesis-selection operation. The value of $N$ used in the reported experiments is provided in the Experimental Setup section.

For compactness, let $x_{t-1}=(b,h_{t-1},f_{t-1})$. This iterative generation process approximates the joint conditional distribution:

\begin{equation}
\label{eq:3}
\begin{split}
& P\!\left(h_{1:T},S^{(1:T)}\mid b,I\right)\\
&\quad =
\prod_{t=1}^{T}
P\!\left(S^{(t)}\mid x_{t-1},I^{(t)}\right)\\
&\qquad\quad \times
P\!\left(h_t\mid x_{t-1},S^{(t)}\right).
\end{split}
\end{equation}
where $h_{1:T}=(h_1,\ldots,h_T)$ and $S^{(1:T)}=(S^{(1)},\ldots,S^{(T)})$. The first conditional term represents inspiration selection from the dynamically retrieved article set $I^{(t)}$, whereas the second represents hypothesis generation or refinement. We set $h_0=f_0=\emptyset$ for the initial round.

This factorization represents the complete refinement trajectory as a sequence of executable retrieval, inspiration-selection, feedback, and hypothesis-generation steps. Algorithm~\ref{alg:hypothesis} describes one candidate refinement trajectory. The framework applies this process to $N$ candidate trajectories, producing $H^{(T)}=\{h_j^{(T)}\}_{j=1}^{N}$, from which the final hypothesis $h^*$ is selected. The number of refinement rounds $T$ is fixed by the predefined temporal partition $\{\Delta_t\}_{t=1}^{T}$ and is not learned by the model. The temporal intervals and the value of $T$ used in the experiments are reported in the Experimental Setup section.

\begin{algorithm}[t]
\caption{Temporal Hypothesis Refinement for One Candidate Trajectory}
\label{alg:hypothesis}
\textbf{Input}: Background $b$, temporal intervals $\{\Delta_t\}_{t=1}^{T}$\\
\textbf{Output}: Refined candidate hypothesis $h_T$

\begin{algorithmic}[1]
\STATE Extract chemically relevant keywords from $b$
\STATE $q_1\leftarrow
\operatorname{Query Construction}(\text{extracted keywords})$
\STATE $I^{(1)}\leftarrow
\operatorname{Retrieve}(q_1,\Delta_1)$
\STATE $S^{(1)}\leftarrow
\operatorname{Inspiration Selection}(I^{(1)}\mid b)$
\STATE $h_1\leftarrow\operatorname{Generation}(b,S^{(1)})$
\FOR{$t=2$ to $T$}
    \STATE $f_{t-1}\leftarrow
    \operatorname{Feedback}(b,h_{t-1},S^{(t-1)})$
    \STATE Extract keywords from $b$, $h_{t-1}$, and $f_{t-1}$
    \STATE $q_t\leftarrow
    \operatorname{Query Construction}(\text{extracted keywords})$
    \STATE $I^{(t)}\leftarrow
    \operatorname{Retrieve}(q_t,\Delta_t)$
    \STATE $S^{(t)}\leftarrow
    \operatorname{Inspiration Selection}(I^{(t)}
    \mid b,h_{t-1},f_{t-1})$
    \STATE $h_t\leftarrow
    \operatorname{Generation}(b,h_{t-1},f_{t-1},S^{(t)})$
\ENDFOR
\STATE \textbf{return} $h_T$
\end{algorithmic}
\end{algorithm}

\subsection{LLM-Guided Literature Retrieval}

A key component of the MAIL pipeline is a dynamic, LLM-guided literature retrieval module. Instead of relying on static keyword lists, hand-crafted queries, or predefined corpora, we use a prompt-based approach to extract high-precision, domain-specific keywords directly from the background knowledge.

\subsubsection{Keyword Extraction.}
Given the background information $b$, we first prompt the language model to summarize and structure it into interpretable components, including cause-effect relationships and problem-solution-outcome representations. Using this structured output, we then prompt the model to extract chemically relevant search terms through a task-specific instruction. The prompt is designed to guide the model in extracting chemically grounded concepts suitable for a literature search. These include both explicit keywords and latent domain-specific entities inferred from context, such as evaluation metrics, functional roles, or material classes. By reasoning beyond explicitly stated phrases, the model can identify meaningful search tokens even if not directly mentioned. Generic or ambiguous terms are filtered out to ensure relevance and precision.

\subsubsection{Query Construction.}
Instead of manually composing queries, the model uses the extracted keywords to generate structured Boolean search strings. These queries are designed to be compatible with academic search platforms. The model constructs queries by logically grouping domain-specific terms to narrow the search scope and improve precision.

Each generated query is executed across three major scientific literature databases: PubMed, CrossRef, and Semantic Scholar. A publication date filter is applied to restrict results to the specific time slice, ensuring that only temporally prior knowledge is retrieved at each refinement step. Retrieved results from all three sources are aggregated, de-duplicated, and merged into a unified set of candidate inspiration papers $I^{(t)}$.

\subsection{Incremental Refinement and Self-Evaluation}

After retrieving candidate papers for the current time slice $I^{(t)}$, rather than relying on embedding similarity or static filtering, the model is asked to identify which papers offer the most conceptually or mechanistically useful insights for background $b$. For each selected paper, it generates a structured explanation that includes the title, a summary of the relevant idea or finding, and a justification for how that content could meaningfully direct the current hypothesis. This results in an interpretable and context-sensitive inspiration set $S^{(t)}$.

Each element $s^{(t)}\in S^{(t)}$ is represented as a structured record:
\[
s^{(t)}=
(\text{title},\text{summary},\text{justification}).
\]
The summary identifies the finding relevant to the current problem, while the justification explains how that finding could support or improve the current hypothesis. These structured records are passed directly to the hypothesis-generation prompt.

At $t = 1$, the language model generates an initial hypothesis $h_1$ using only the background $b$ and the inspiration set $S^{(1)}$. The prompt at this stage is designed to synthesize known information into a plausible and mechanistically grounded research direction based solely on prior literature.

Before each refinement round $t>1$, the model evaluates the preceding hypothesis $h_{t-1}$ and produces structured feedback $f_{t-1}$. This evaluation follows four criteria: validity, novelty, significance, and potential. The model assesses whether the hypothesis is scientifically plausible and experimentally feasible (validity), introduces ideas beyond established literature (novelty), could meaningfully advance the field (significance), and could become more impactful through further methodological development (potential). Rather than providing a generic critique, the feedback identifies specific weaknesses, such as vague design steps, implausible mechanisms, insufficient differentiation from prior work, or missing experimental details, and provides actionable suggestions for the next refinement round.

For $t > 1$, the previous hypothesis ($h_{t-1}$) and the feedback ($f_{t-1}$) are leveraged to guide the subsequent reasoning round. This feedback plays a dual role: it informs the selection of more relevant inspiration materials ($S^{(t)}$) for the next round and critically conditions the generation of the updated hypothesis. Specifically, hypothesis generation for $t > 1$ is conditioned on four components: the background ($b$), the previous hypothesis ($h_{t-1}$), the feedback on previous hypothesis ($f_{t-1}$), and the current inspiration set ($S^{(t)}$). The model is explicitly instructed to preserve strong elements of the prior hypothesis, integrate new insights from inspiration papers, and revise based on the feedback. This iterative process allows the hypothesis to evolve gradually over consecutive intervals. 

\subsection{Prompt Adaptation and Hypothesis Selection}

The final stage of the MAIL method focuses on identifying the most promising hypothesis through adaptive prompting, memory-driven evaluation, and iterative refinement. Unlike single-pass generation approaches, we treat the hypothesis selection as a multi-round reasoning task, where each iteration benefits from accumulated context and structured self-evaluation.

Here, memory-driven refers to the use of the accumulated memory summary from preceding rounds to guide subsequent inspiration selection, hypothesis refinement, and final selection. In contrast, literature-grounded refers to the use of retrieved scientific publications as external evidence during hypothesis generation.

To simulate long-term memory and maintain conceptual consistency, we introduce a compressed contextual representation inspired by in-context learning strategies such as Multi-modal Contextual Vectors (MCVs) \citep{huang2024multimodal, liu2024context}. After each round, the method summarizes prior inspiration materials into a structured paragraph capturing relevant mechanisms, materials, and scientific themes. This summary is added to subsequent prompts, allowing the LLM to stay grounded in earlier insights without overwhelming the input context. The same memory structure is reused during final selection.

To further guide the generation process, we implement an adaptive evaluation and feedback loop on the selection prompt. If the currently selected hypothesis exhibits weaknesses, such as redundancy, vague mechanisms, or lack of novelty, the model is prompted to revise the selection criteria. The revised prompt emphasizes weaknesses, highlights areas for improvement, and directs the next round of generation toward more refined outputs.

This iterative prompting strategy, combined with latent memory compression and adaptive self-assessment, enables the model to refine its decision-making logic over time. The result is not a single prediction, but a product of cumulative reasoning, mimicking the way human researchers revisit and reshape their ideas through iterative exploration. By retaining conceptual continuity across refinement stages, MAIL ensures that promising but underdeveloped ideas are not discarded prematurely, but instead nurtured into coherent and scientifically plausible hypotheses. The full text of the prompts used at each stage of this framework can be found in Appendix \ref{prompts_details}.

\section{Evaluation and Results}

\subsection{Dataset}
We evaluate our method on the public TOMATO-Chem benchmark \citep{yang2024moose}, which assesses LLMs on hypothesis generation in chemistry and materials science using 51 papers spanning areas such as organic and polymer chemistry. This dataset simulates a co-pilot setting in which a researcher provides a question and a brief survey, allowing the assessment of overall performance. 

Similar to recent efforts that introduce specialized benchmarks to rigorously test LLM inference capabilities \citep{van2025new}, to further test reasoning depth in our domain, we curated a high-novelty nature/science challenge dataset (HN-NS) comprising 50 research chemistry papers recently published in Nature and Science. This dataset targets frontier topics in Electrochemistry, Biocatalysis, and Materials Science that are inherently challenging for literature-trained models for three reasons. First, these underexplored areas often suffer from conceptual sparsity, where a limited number of established relationships restricts the model’s ability to form novel conceptual links. Second, these scenarios frequently involve evolving scientific theories or modern experimental evidence that deviates from established paradigms, requiring the model to reason beyond well-trodden patterns. Finally, this set includes discoveries inspired by implicit information, such as side reactions or secondary observations, that receive limited emphasis in standard abstracts and summaries. 

Same as the TOMATO-Chem benchmark, for each paper, expert annotators decomposed the content into structured components, including: the research background, which consists of a background question and a background survey, two inspiration paper titles along with their justification for serving as an inspiration, the target hypothesis, and the main points of the target hypothesis. Importantly, the background and inspiration components are carefully collected to avoid leakage, ensuring that models must reason forward from a limited context to reconstruct the original idea.

\begin{table}[htbp] 
\caption{Evaluation rubric for research hypotheses in chemistry. Each dimension is scored on a 0–2 scale based on the provided definitions and scoring tips.}\label{tab:evaluation_rubric}
\centering
\small
\begin{tabular*}{\tblwidth}{@{\extracolsep{\fill}} p{3cm} p{4cm} p{7.4cm} @{}}
\toprule
\textbf{Dimension} & \textbf{Definition} & \textbf{Scoring Tips} \\
\midrule
Chemical Plausibility & Is the hypothesis chemically sound and consistent with known mechanisms or principles? & 
0 = Contradicts known chemistry or is implausible; 
1 = Speculative but not ruled out by current understanding; 
2 = Strongly consistent with known mechanisms or literature precedent. \\
\addlinespace
Physical Feasibility & Can the hypothesis be tested using accessible methods or tools? & 
0 = Requires inaccessible, unscalable, or unknown techniques; 
1 = Needs careful optimization, special reagents, or advanced equipment; 
2 = Readily testable using standard lab procedures. \\
\addlinespace
Novelty – Technical & Does it introduce new catalysts, substrates, experimental conditions, or materials? & 
0 = Uses standard techniques or well-known combinations; 
1 = Applies known tools in a slightly modified or new way; 
2 = Introduces a new catalyst, scaffold, platform, or methodology. \\
\addlinespace
Novelty – Conceptual & Does it introduce a new chemical concept, reaction logic, or mechanistic rationale? & 
0 = Based entirely on established chemical concepts and reaction logic, without reinterpretation or mechanistic novelty; 
1 = Recombines familiar mechanistic ideas or reaction types creatively or unexpectedly, but without introducing fundamentally new logic; 
2 = Proposes a fundamentally new chemical concept, novel reaction logic, or mechanistic rationale that could reshape how related systems are understood. \\
\addlinespace
Completeness & Is the hypothesis specific and testable as stated? & 
0 = Vague idea without concrete components or outcomes; 
1 = Some elements are defined, but others are missing; 
2 = Clearly specifies inputs, conditions, and measurable outputs. \\ 
\addlinespace
Scientific Impact & Could it change how chemists think, plan, or execute reactions? & 
0 = Minor or incremental advancement; 
1 = Could improve performance or solve a niche problem; 
2 = May shift strategy, or design at a broader level. \\
\addlinespace 
Broader Utility & Does the idea have wide potential across substrates, scales, or industries? & 
0 = Narrow use (e.g., one substrate or condition); 
1 = Niche relevance (e.g., specific academic or materials context); 
2 = Broad synthetic utility or clear industrial/pharmaceutical relevance. \\
\bottomrule
\end{tabular*}
\end{table} 

\subsection{Evaluation Approach}
Direct prospective evaluation of generated chemical hypotheses would require extensive experimental validation and is not feasible at benchmark scale. We therefore adopt a retrospective temporal evaluation as a practical proxy for prospective hypothesis generation. For each target, the model is restricted to the background and literature available before the target discovery, while the subsequently published paper provides a peer-reviewed and experimentally supported reference hypothesis. Successful recovery therefore indicates that the framework can synthesize, from temporally prior evidence, an idea that was historically novel and scientifically consequential at the time of publication. Nevertheless, MIOS and MPOS quantify recovery of that reference idea rather than independent originality of every generated alternative.

To assess the effectiveness of MAIL, we adopt both automated and expert evaluation protocols, comparing our method's outputs against ground-truth hypotheses from the benchmark. Following the evaluation metrics and protocol in \citep{yang2024moose}, the generated hypothesis is evaluated using the main idea overlap score (MIOS) and main points overlap score (MPOS). For the principal comparisons, the baseline methods were evaluated using their released implementations, and all methods were run during the same experimental period using the same GPT-4o configuration and temporal cutoff. Each method, including every reported baseline, was independently executed ten times for each background, producing one finalized hypothesis per run. Within each MAIL run, candidate reduction and final selection are performed automatically using the predefined selection prompts, without access to the target hypothesis or the MIOS and MPOS scores. We report the mean across the ten finalized outputs as the primary statistic. Following the benchmark reporting convention, we additionally report Top, defined as the post-hoc maximum score among the same ten outputs. This score is used only for evaluation and is never fed back into hypothesis generation, refinement, or selection. The automatic evaluator is invoked only after hypothesis generation and final selection have been completed. The target hypothesis, MIOS and MPOS evaluation prompts, and resulting scores are never provided to the retrieval, feedback, refinement, or selection stages. Therefore, MAIL is not iteratively optimized against the automatic evaluator. Moreover, MIOS explicitly evaluates alignment of the central scientific claim and mechanism while excluding writing quality, length, and stylistic features, whereas MPOS evaluates coverage of the predefined methodological points in the reference hypothesis. All outputs are evaluated without method-identifying information using the same fixed rubrics.

Since both datasets consist only of publications after the underlying LLM's knowledge cutoff (GPT-4o, Oct 2023), they provide a rigorous environment for testing autonomous hypothesis generation without the risk of data contamination. This setting provides a realistic and rigorous analysis for benchmarking automated scientific reasoning systems in the generation of chemical hypotheses. 

To assess the domain alignment and scientific utility of the generated hypotheses, we conducted a structured expert evaluation. Detailed definitions and scoring guidelines can be found in Table~\ref{tab:evaluation_rubric}. A PhD-level organic chemist reviewed hypotheses generated from the organic chemistry subset of the benchmark dataset, evaluating outputs generated by both the MAIL and MOOSE-Chem methods. The evaluation was conducted in a double-blind fashion to avoid bias with respect to the source method. These seven criteria included: chemical plausibility, physical feasibility, novelty (conceptual and technical), completeness, scientific impact, and broader utility. This multifaceted rubric ensures a balanced assessment that goes beyond single-criterion evaluation, enabling analysis of trade-offs between creativity, feasibility, and potential impact. 

\subsection{Experimental setup}
All generations were performed using the GPT-4o API with a temperature of 0.7 to balance novelty and coherence. For each background question, the model empirically executes a three-round iterative refinement process. Literature retrieval integrates PubMed, Semantic Scholar, and CrossRef APIs, which are filtered and grouped based on publication year (pre-2015, 2015–2020, 2021-2023) to simulate temporal reasoning following literature partitioning strategies similar to \citep{akujuobi2024link}. All literature searches were restricted to records dated no later than the end of 2023. Retrieved records matching a target paper or an alternate version of that paper by title or persistent identifier were excluded. Closely related prior studies were retained because they constitute legitimate temporally available evidence rather than target leakage. Abstracts were automatically preprocessed to remove duplicates based on semantic title similarity and cleaned for formatting inconsistencies. Each MAIL execution consists of three refinement rounds and generates 8-10 candidate hypotheses per round, corresponding to 24-30 candidate outputs and approximately 35-40 LLM calls per execution. Retrieval uses a depth of five records per query, from which two inspiration papers are selected in each round. PubMed, CrossRef, and Semantic Scholar are queried in every round, resulting in nine source-level retrieval requests per execution, excluding pagination and failed-request retries. Candidate hypotheses are generated through separate calls. Failed requests are retried up to three times only for network or rate-limit errors; no quality-dependent or evaluator-guided regeneration is performed. The comparisons in this study evaluate complete methods under their documented configurations rather than under identical internal search structures. MOOSE-Chem is also a multi-round candidate-search framework that uses beam search, hypothesis ranking, and evolutionary refinement. Simpler baselines require fewer inference operations. Consequently, the reported results compare end-to-end hypothesis quality and should not be interpreted as computational efficiency or performance per model call, token, runtime, or API cost.

\begin{table}[htbp]
\caption{Comparison of MAIL and baseline methods on two MIOS and MPOS on the TOMATO-Chem dataset. \textbf{Bold} values indicate the best performance, and \underline{underlined} values indicate the second-best performance.}\label{tab:combined_mios_mpos}
\centering
\begin{tabular*}{\tblwidth}{@{\extracolsep{\fill}} lcccc @{}}
\toprule
Method & Top MIOS & Avg MIOS & Top MPOS & Avg MPOS \\
\midrule
MOOSE \citep{yang2023large} & 3.18 & 2.52 & 1.94 & 1.85 \\
Zero-shot Proposer \citep{qi2023large} & 2.49 & 2.12 & 1.74 & 1.53 \\
SciMON \citep{wang2024scimon} & 2.71 & 2.17 & 1.91 & 1.73 \\
NOVA \citep{hu2025nova} & 2.33 & 2.04 & 1.59 & 1.28 \\
ResearchBench \citep{liu2025researchbench} & 3.41 & 2.84 & 2.30 & 1.87 \\
MOOSE-Chem \citep{yang2024moose} & \underline{4.01} & \underline{3.15} & \underline{3.56} & \underline{2.14} \\
\textbf{MAIL (Ours)} & \textbf{4.27} & \textbf{3.69} & \textbf{3.91} & \textbf{2.38} \\
\bottomrule
\end{tabular*}
\end{table}

Table~\ref{tab:combined_mios_mpos} reports the comparison results with six state-of-the-art (SOTA) methods (i.e., MOOSE \citealp{yang2023large}, Zero-shot Proposer \citep{qi2023large}, SciMON \citep{wang2024scimon}, MOOSE-Chem \citep{yang2024moose}, NOVA \citep{hu2025nova}, and ResearchBench \citep{liu2025researchbench}) designed for scientific hypothesis generation. In the principal comparisons, the same underlying generator and automatic evaluator are used for all methods. For the open-weight model experiments, MAIL and MOOSE-Chem are compared under the same generator within each model family; these results are therefore interpreted as within-generator comparisons rather than absolute comparisons across model families.

The results demonstrate MAIL's potential to synthesize historically novel research directions from temporally prior literature. MOOSE-Chem \citep{yang2024moose}, as a chemistry version of MOOSE \citep{yang2023large}, delivers high results due to domain-specific tuning but relies on a static inspiration corpus. This limits adaptability to evolving research directions. The improvement of MAIL over MOOSE-Chem suggests that our design, particularly dynamic context-aware retrieval and multi-stage refinement, improves the recovery of historically novel target ideas and their principal mechanisms. Zero-shot Proposer \citep{qi2023large} underperforms due to its lack of refinement and domain adaptation. The generated hypotheses remain inclusive and underspecified, resulting in lower MIOS and MPOS scores. SciMON \citep{wang2024scimon} produces outputs that are comparatively vague and weakly aligned with experimental structures. The retrieval method is less likely to consistently surface mechanistically grounded ideas, limiting its performance. NOVA \citep{hu2025nova} emphasizes idea diversity over mechanistic validity. Its hypotheses lack the structural and physical grounding needed for chemistry, which results in lower scores on both evaluation metrics. ResearchBench \citep{liu2025researchbench} shows competitive results overall but lacks temporal refinement and domain-specific feedback, constraining its ability to develop hypotheses that evolve with emerging literature. 

\begin{table}[htbp]
\caption{Ablation study. The average of MIOS and MPOS over the benchmark. Each variant removes one key component of the full method. \textbf{Bold} values indicate the best performance, and \underline{underlined} values indicate the second-best performance}\label{tab:ablation_study}
\centering
\begin{tabular*}{\tblwidth}{@{\extracolsep{\fill}} c c c c c c @{}}
\toprule
\shortstack{Iterative \\ Refinement} &
\shortstack{Feedback \\ Reasoning} &
\shortstack{Context-Aware \\ Prompts} &
\shortstack{Prompt \\ Optimization} &
\shortstack{Avg MIOS} &
\shortstack{Avg MPOS} \\
\midrule
\includegraphics[width=0.03\columnwidth]{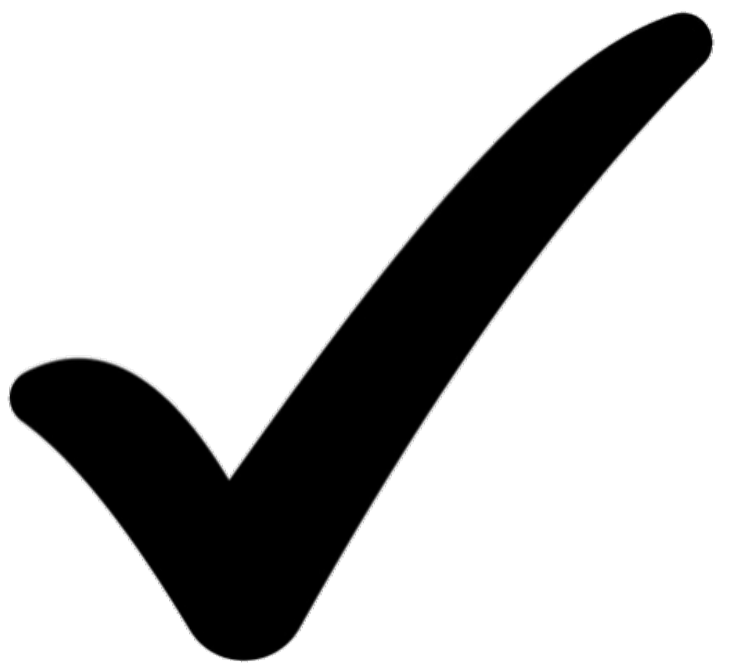} & \includegraphics[width=0.03\columnwidth]{checkmark} & \includegraphics[width=0.03\columnwidth]{checkmark} & \includegraphics[width=0.03\columnwidth]{checkmark} & \textbf{3.69} & \textbf{2.38} \\
$\times$ & \includegraphics[width=0.03\columnwidth]{checkmark} & \includegraphics[width=0.03\columnwidth]{checkmark} & \includegraphics[width=0.03\columnwidth]{checkmark} & 2.59 & 1.74 \\
\includegraphics[width=0.03\columnwidth]{checkmark} & $\times$ & \includegraphics[width=0.03\columnwidth]{checkmark} & \includegraphics[width=0.03\columnwidth]{checkmark} & \underline{3.03} & \underline{2.07} \\
\includegraphics[width=0.03\columnwidth]{checkmark} & \includegraphics[width=0.03\columnwidth]{checkmark} & $\times$ & \includegraphics[width=0.03\columnwidth]{checkmark} & 2.35 & 1.46 \\
\includegraphics[width=0.03\columnwidth]{checkmark} & \includegraphics[width=0.03\columnwidth]{checkmark} & \includegraphics[width=0.03\columnwidth]{checkmark} & $\times$ & 2.29 & 1.51 \\
\bottomrule
\end{tabular*}
\end{table}

\begin{table}[htbp]
\caption{Expert evaluation results on TOMATO-Chem dataset. Scores range from 0 to 2 per criterion. \textbf{Bold} values indicate the best performance.}\label{tab:expert_eval}
\centering
\scriptsize
\setlength{\tabcolsep}{4pt}
\begin{tabular*}{\tblwidth}{@{\extracolsep{\fill}} l c c c c c c c c @{}}
\toprule
 &
\shortstack{Chemical \\ Plausibility} &
\shortstack{Physical \\ Feasibility} &
\shortstack{Novelty \\ (Technical)} &
\shortstack{Novelty \\ (Conceptual)} &
\shortstack{Completeness} &
\shortstack{Scientific \\ Impact} &
\shortstack{Broader \\ Utility} &
\shortstack{Total \\ Score} \\
\midrule
MOOSE-Chem & 1.36 & 1.20 & \textbf{1.64} & 1.28 & 1.68 & 1.68 & 1.48 & 10.32 \\
MAIL (Ours)       & \textbf{1.44} & \textbf{1.32} & 1.56 & \textbf{1.36} & \textbf{1.72} & \textbf{1.76} & \textbf{1.64} & \textbf{10.80} \\
\bottomrule
\end{tabular*}
\end{table}

\begin{table}[htbp]
\caption{Experiments on the effect of the number of retrieved papers ($K$) per query on TOMATO-Chem dataset. \textbf{Bold} values indicate the best performance, and \underline{underlined} values indicate the second-best performance.}\label{tab:num_retrieved}
\centering
\small
\begin{tabular*}{\tblwidth}{@{\extracolsep{\fill}} ccccc @{}}
\toprule
\shortstack{\textbf{Num. Papers} \\ \textbf{(K) per query}} & \textbf{Top MIOS} & \textbf{Avg MIOS} & \textbf{Top MPOS} & \textbf{Avg MPOS} \\
\midrule
1 & 3.64 & 2.95 & 3.15 & 1.93 \\
3 & \underline{3.91} & \underline{3.27} & \underline{3.47} & \underline{2.05} \\
5 & \textbf{4.27} & \textbf{3.69} & \textbf{3.91} & \textbf{2.38} \\
10  & 3.36 & 3.01 & 2.90 & 1.72 \\
\bottomrule
\end{tabular*}
\end{table}

\begin{figure}[t]
\centering
\includegraphics[width=0.7\columnwidth]{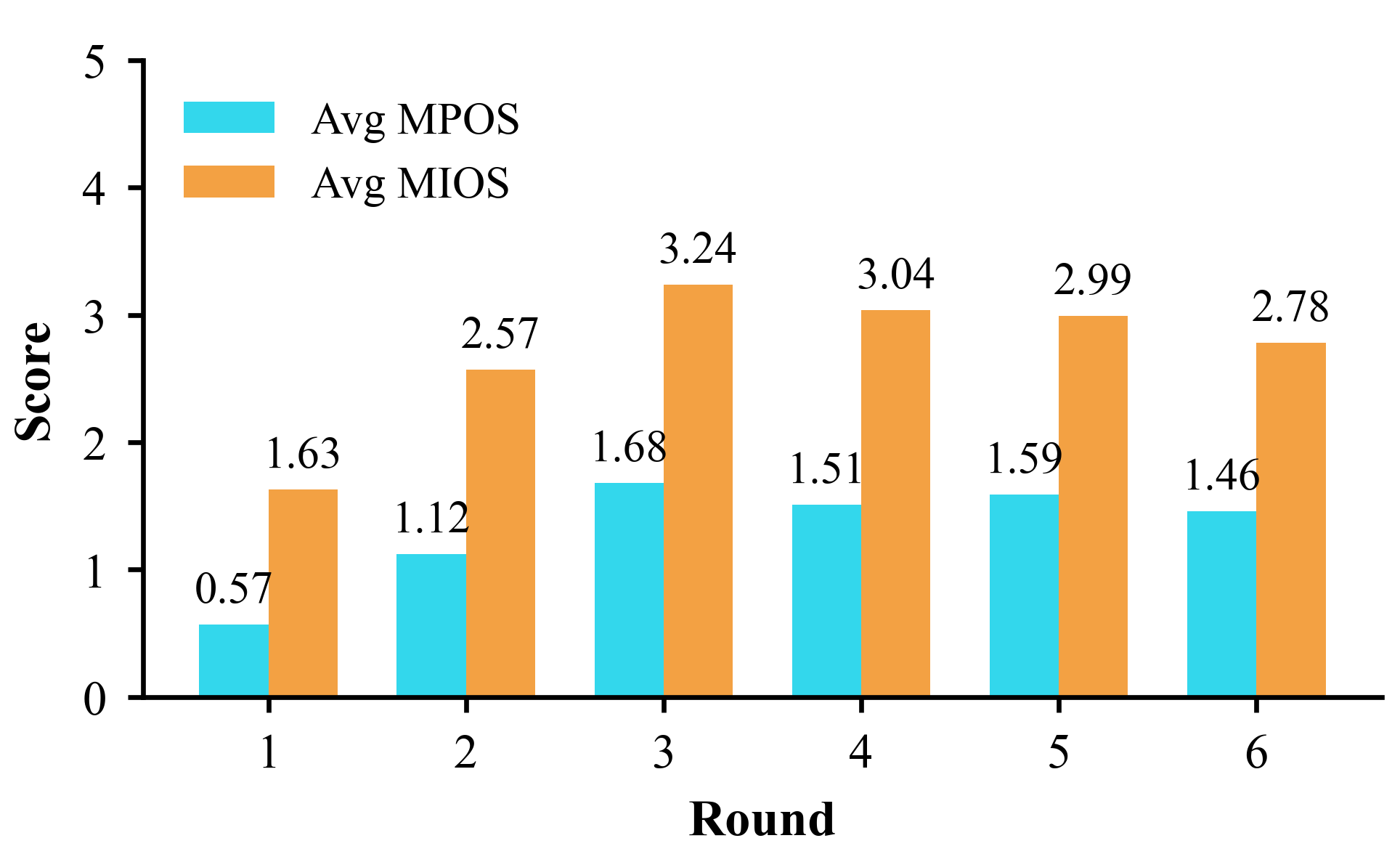}
\caption{
Performance progression across iterative refinement rounds. The final column represents the
average scores of the selected hypotheses.
}
\label{fig:progression}
\end{figure}

\subsection{Ablation Study and Iterative Refinement}

We conduct an ablation study by systematically disabling key modules: iterative refinement, feedback reasoning, context-aware retrieval, and prompt optimization. As shown in Table~\ref{tab:ablation_study},
removing iterative refinement results in a noticeable drop in both MIOS and MPOS, confirming the value of evolving hypotheses over multiple rounds. This shows that complex scientific ideas are rarely formed in a single pass but rather emerge through gradual improvement and accumulated reasoning. 
Excluding context-aware retrieval or prompt optimization results in a performance drop, suggesting that these components play a central role in grounding hypothesis generation in relevant prior knowledge and maintaining conceptual consistency. Without timely, context-specific inspirations or dynamically adapted prompts, the model struggles to align its outputs with the background problem or produce mechanistically rich ideas, indicating the importance of both information quality and prompt conditioning in scientific ideation.

We also study the effect of iterative refinement. Fig.~\ref{fig:progression} shows the progression across rounds. Both MIOS and MPOS improve steadily from round 1 to round 3, with the final selected hypotheses achieving the highest scores. This analysis confirms that refinement over time, through new literature, accumulated feedback, and evolving hypotheses, enables the method to generate more coherent, novel, and well-structured outputs. These results confirm model hypothesis generation as a staged reasoning process rather than a single-pass generation. A detailed, step-by-step example of this refinement progression is provided in Appendix \ref{case_study_example}.

\begin{figure}[t]
  \centering
  \includegraphics[width=0.7\textwidth]{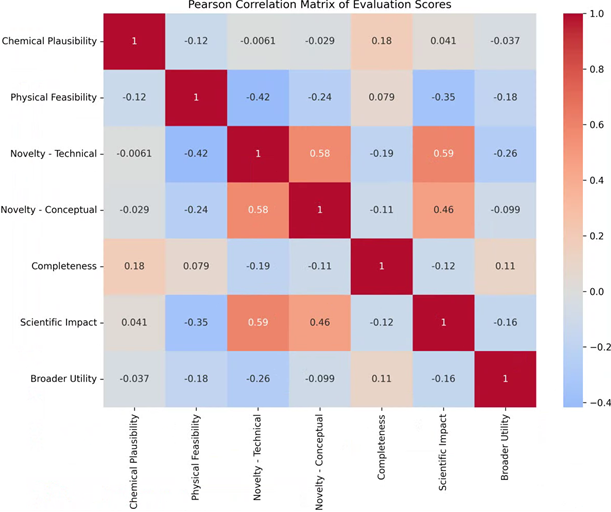}
  \caption{Pearson correlation matrix of expert evaluation scores for generated hypotheses on TOMATO-Chem dataset. 
Cells are color-coded from blue (negative correlation) to red (positive correlation), 
with the intensity indicating the magnitude of the association.}
  \label{fig:perf_quality_report}
\end{figure}

\section{Expert Evaluation}

We conducted structured expert evaluations on 25 hypotheses from each dataset, totaling 50 reviews by a PhD-level organic chemist. On TOMATO-Chem (Table~\ref{tab:expert_eval}), MAIL achieves a higher overall expert-evaluation score than MOOSE-Chem (10.80 versus 10.32), with stronger scores in chemical plausibility, physical feasibility, conceptual novelty, completeness, scientific impact, and broader utility. MOOSE-Chem receives a higher technical-novelty score (1.64 versus 1.56), whereas MAIL receives a higher conceptual-novelty score (1.36 versus 1.28).

Additionally, we explored associations among evaluation criteria using Pearson correlation coefficients ($N = 25$). Given the limited sample size ($N=25$ per dataset) and the ordinal 0--2 scoring scale, these correlation analyses are considered exploratory. Approximate 95\% confidence intervals were calculated using Fisher's $r$-to-$z$ transformation, and two-sided $p$-values were obtained using the standard Pearson correlation test. To account for the 21 unique pairwise comparisons among seven criteria, the resulting $p$-values were adjusted separately for each dataset using the Benjamini-Hochberg procedure. As shown in Fig.~\ref{fig:perf_quality_report}, for the TOMATO-Chem dataset, technical novelty showed exploratory positive associations with conceptual novelty ($r=0.58$, 95\% CI $[0.24,0.79]$, $q=0.025$) and scientific impact ($r=0.59$, 95\% CI $[0.25,0.80]$, $q=0.025$). These associations remained below the Benjamini--Hochberg-adjusted threshold. The remaining correlations, including the negative relationship between physical feasibility and technical novelty, did not remain statistically significant after correction and are therefore interpreted descriptively.

\begin{table}[htbp]
\caption{Experiments on the effect of different temperature values on hypothesis generation performance. \textbf{Bold} values indicate the best performance, and \underline{underlined} values indicate the second-best performance.}\label{tab:temperature}
\centering
\small
\begin{tabular*}{\tblwidth}{@{\extracolsep{\fill}} ccccc @{}}
\toprule
\textbf{Temperature} & \textbf{Top MIOS} & \textbf{Avg MIOS} & \textbf{Top MPOS} & \textbf{Avg MPOS} \\
\midrule
0.0 & 3.85 & 1.85 & 3.01 & 1.93 \\
0.3 & 3.88 & 3.34 & 3.27 & 2.02 \\
0.5 & 4.04 & 3.32 & \textbf{3.93} & \underline{2.28} \\
0.7  & \underline{4.27} & \textbf{3.69} & \underline{3.91} & \textbf{2.38} \\
1.0 & \textbf{4.29} & \underline{3.48} & 3.35 & 2.09 \\
\bottomrule
\end{tabular*}
\end{table}

\section{Parameter Sensitivity Analysis}
Table~\ref{tab:num_retrieved} evaluates the effect of the number of retrieved papers ($K$) per query. The results indicate that a moderate number of retrieved papers leads to better hypothesis quality, while both too few and too many inspirations reduce performance. This confirms that MAIL benefits from a controlled diversity of inspirations without overloading the model with irrelevant context. 

To examine the robustness of our conclusions concerning method parameters, we additionally analyze the impact of the LLM temperature in the MAIL framework. All evaluations use GPT-4o (trained on data up to October 2023). Performance is measured using MIOS (Main Idea Overlap Score) and MPOS (Main Points Overlap Score). Table~\ref{tab:temperature} reports results for varying the temperature parameter, illustrating the evaluation setup in the MAIL method. The results show that intermediate temperature values yield a favorable balance between novelty and specificity, supporting our choice of hyperparameters.

\begin{table}[htbp]
\caption{Performance evaluation on HN-NS Dataset. Each value reflects the top-scoring and average performance per method. \textbf{Bold} values indicate the best performance, and \underline{underlined} values indicate the second-best performance.}\label{tab:combined_mios_mpos_2}
\centering
\begin{tabular*}{\tblwidth}{@{\extracolsep{\fill}} lcccc @{}}
\toprule
Method & Top MIOS & Avg MIOS & Top MPOS & Avg MPOS \\
\midrule
MOOSE \citep{yang2023large} & 2.41 & 1.78 & 1.62 & 1.55 \\
Zero-shot Proposer \citep{qi2023large} & 1.88 & 1.50 & 1.39 & 1.34 \\
SciMON \citep{wang2024scimon} & 2.05 & 1.49 & 1.53 & 1.38 \\
NOVA \citep{hu2025nova} & 1.76 & 1.44 & 1.27 & 1.12 \\
ResearchBench \citep{liu2025researchbench} & 2.78 & 2.11 & 1.97 & 1.64 \\
MOOSE-Chem \citep{yang2024moose} & \underline{3.03} & \underline{2.23} & \underline{2.85} & \underline{1.88} \\
\textbf{MAIL (Ours)} & \textbf{3.23} & \textbf{2.61} & \textbf{3.13} & \textbf{2.09} \\
\bottomrule
\end{tabular*}
\end{table}

\begin{table}[htbp]
\caption{Expert evaluation results on HN-NS Dataset. Scores range from 0 to 2 per criterion. \textbf{Bold} values indicate the best performance. }\label{tab:expert_eval_2}
\centering
\scriptsize
\setlength{\tabcolsep}{4pt}
\begin{tabular*}{\tblwidth}{@{\extracolsep{\fill}} l c c c c c c c c @{}}
\toprule
 &
\shortstack{Chemical \\ Plausibility} &
\shortstack{Physical \\ Feasibility} &
\shortstack{Novelty \\ (Technical)} &
\shortstack{Novelty \\ (Conceptual)} &
\shortstack{Completeness} &
\shortstack{Scientific \\ Impact} &
\shortstack{Broader \\ Utility} &
\shortstack{Total \\ Score} \\
\midrule
MOOSE-Chem & 1.21 & 1.08 & 1.08 & 1.00 & \textbf{1.33} & 1.42 & 1.21 & 8.33 \\
MAIL (Ours)       & \textbf{1.42} & 1.08 & \textbf{1.17} & 1.00 & 1.25 & \textbf{1.46} & \textbf{1.25} & \textbf{8.63} \\
\bottomrule
\end{tabular*}
\end{table}

\section{Experiments on HN-NS Dataset}
To validate the generalizability, we also conduct automated and human evaluations on the HN-NS dataset. The evaluation on our HN-NS dataset (Table~\ref{tab:combined_mios_mpos_2}) demonstrates that MAIL maintains a robust Main Idea Overlap Score even on a more challenging benchmark. Compared to the primary TOMATO-Chem suite, the new dataset involves a higher degree of conceptual novelty. Reaching these expert-level insights requires the framework to perform complex, non-obvious synthesis across disparate domains.

\begin{figure}[t]
  \centering
  \includegraphics[width=0.7\textwidth]{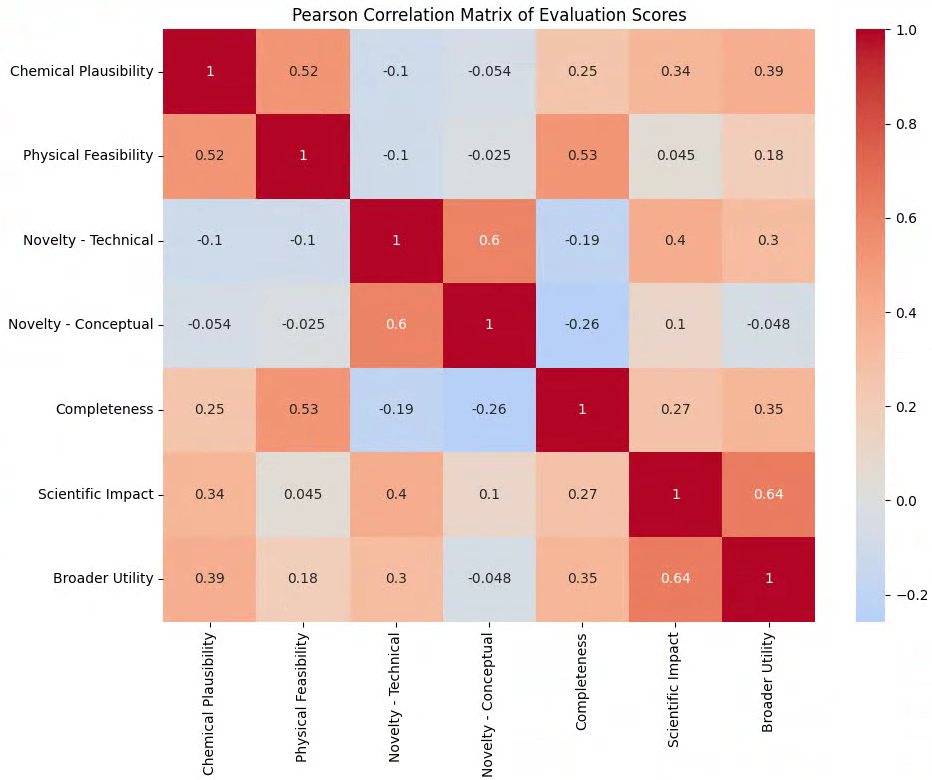}
  \caption{Pearson correlation matrix of expert evaluation scores for generated hypotheses on HN-NS dataset. 
Cells are color-coded from blue (negative correlation) to red (positive correlation), 
with the intensity indicating the magnitude of the association.}
  \label{fig:perf_quality_report_2}
\end{figure}

On expert evaluation (Table~\ref{tab:expert_eval_2}), total scores decreased for all methods due to the increased difficulty of this dataset. These scenarios require a larger inferential distance to bridge the gap between inspiration and target. Despite the higher conceptual complexity, MAIL successfully synthesized non-obvious mechanistic pathways. This indicates that MAIL’s iterative refinement and memory-driven reasoning effectively handle high-novelty frontiers.

Additionally, Fig.~\ref{fig:perf_quality_report_2} shows that physical feasibility maintains a positive correlation with completeness ($r = 0.53$) and scientific impact remains strongly associated with broader utility ($r = 0.64$). Furthermore, technical and conceptual novelty exhibit a stronger mutual correlation ($r = 0.60$) in this setting. For HN-NS, four exploratory associations remained below the adjusted threshold: chemical plausibility with physical feasibility ($r=0.52$, 95\% CI $[0.16,0.76]$, $q=0.040$), physical feasibility with completeness ($r=0.53$, 95\% CI $[0.17,0.77]$, $q=0.040$), technical novelty with conceptual novelty ($r=0.60$, 95\% CI $[0.27,0.80]$, $q=0.016$), and scientific impact with broader utility ($r=0.64$, 95\% CI $[0.33,0.83]$, $q=0.012$).

\section{Impact of Underlying Language Models}

To investigate the generalizability of the MAIL framework across varying foundational capabilities, we evaluated our method using a diverse set of open-weight LLMs. For this analysis, we selected MOOSE-Chem as the comparative baseline, as it demonstrated the most competitive performance among the state-of-the-art methods in our primary evaluations (Tables~\ref{tab:combined_mios_mpos} and \ref{tab:combined_mios_mpos_2}). Specifically, we tested the Qwen2.5 (72B, 32B) \citep{qwen2.5} and Llama-3.1 (70B, 8B) \citep{dubey2024llama} families to observe how model scale impacts the quality of generated hypotheses. Table~\ref{tab:llm_scaling} presents the evaluation on both the TOMATO-Chem and HN-NS datasets.

\begin{table}[htbp]
\caption{Performance evaluation of MAIL and MOOSE-Chem across varying LLMs on the TOMATO-Chem and HN-NS datasets. \textbf{Bold} values indicate the best performance for a given model and dataset.}
\label{tab:llm_scaling}
\centering
\small
\begin{tabular*}{\tblwidth}{@{\extracolsep{\fill}} l l c c c c @{}}
\toprule
& & \multicolumn{2}{c}{\textbf{TOMATO-Chem Dataset}} & \multicolumn{2}{c}{\textbf{HN-NS Dataset}} \\
\cmidrule{3-4} \cmidrule{5-6}
\textbf{Method} & \textbf{Model} & \textbf{Avg MIOS} & \textbf{Avg MPOS} & \textbf{Avg MIOS} & \textbf{Avg MPOS} \\
\midrule
MAIL (Ours) & GPT-4o (Baseline) & \textbf{3.69} & \textbf{2.38} & \textbf{2.61} & \textbf{2.09} \\
MOOSE-Chem  & GPT-4o (Baseline) & 3.15 & 2.14 & 2.23 & 1.88 \\
\addlinespace
MAIL (Ours) & Qwen2.5-72B & \textbf{3.61} & \textbf{2.32} & \textbf{2.55} & \textbf{2.04} \\
MOOSE-Chem  & Qwen2.5-72B & 3.04 & 2.11 & 2.20 & 1.86 \\
\addlinespace
MAIL (Ours) & Llama-3.1-70B & \textbf{3.56} & \textbf{2.27} & \textbf{2.51} & \textbf{2.00} \\
MOOSE-Chem  & Llama-3.1-70B & 3.07 & 2.08 & 2.18 & 1.82 \\
\addlinespace
MAIL (Ours) & Qwen2.5-32B & \textbf{3.30} & \textbf{2.16} & \textbf{2.35} & \textbf{1.91} \\
MOOSE-Chem  & Qwen2.5-32B & 2.95 & 1.99 & 2.10 & 1.78 \\
\addlinespace
MAIL (Ours) & Llama-3.1-8B & \textbf{2.55} & \textbf{1.79} & \textbf{1.82} & \textbf{1.58} \\
MOOSE-Chem  & Llama-3.1-8B & 2.35 & 1.68 & 1.68 & 1.50 \\
\bottomrule
\end{tabular*}
\end{table}

The results demonstrate that MAIL consistently outperforms MOOSE-Chem across all tested architectures, affirming that our framework's methodological improvements are model-agnostic. Furthermore, large open-weight models (Qwen2.5-72B and Llama-3.1-70B) achieve highly competitive performance, closely approaching the closed-source GPT-4o baseline on both datasets. 

As expected, hypothesis quality scales positively with model parameter size. A notable observation is the behavior of the performance gap between MAIL and MOOSE-Chem on smaller models (e.g., Llama-3.1-8B). On these smaller architectures, the absolute performance delta shrinks slightly compared to the >70B models. This occurs because the model's fundamental reasoning capabilities bottleneck before the methodological enhancements of MAIL can be fully realized. Nevertheless, even within these constrained environments, the MAIL framework provides a clear and consistent advantage over the single-pass baseline, proving its utility across diverse computational resource settings.

\section{Discussion}

Our findings indicate that Large Language Model (LLM)-based systems, when properly grounded, have the potential to make meaningful contributions to early-stage scientific discovery within complex domains. The evaluation of the MAIL framework on both the TOMATO-Chem and our curated high-novelty (HN-NS) datasets demonstrates strong gains over domain-specific and general-purpose baselines. This confirms that the careful integration of dynamic retrieval, internal feedback, and adaptive prompt design is crucial for producing plausible and mechanistically sound hypotheses.

Despite its promising performance, a few limitations currently constrain the framework. Primarily, while our method shows strong conceptual alignment with ground-truth hypotheses from high-impact literature, the generated ideas have not yet undergone physical validation in a wet-lab setting. Their plausibility currently relies on expert review and literature grounding rather than empirical execution. Furthermore, the iterative refinement process is computationally intensive. Each discovery trajectory requires multiple rounds of LLM reasoning and real-time retrieval from three distinct literature databases, introducing significant API latency and computational cost compared to single-pass generation methods. Additionally, the current evaluation is focused predominantly on organic and materials chemistry, leaving the framework’s utility in more specialized or disparate scientific fields yet to be established. Finally, the autonomous nature of the system presents inherent dual-use risks, as the model could theoretically be misused to generate hypotheses involving hazardous chemical substances.

\section{Future Work}

To address these limitations and expand the framework's capabilities, future work will focus on several key areas. First, we aim to explore deeper alignment with experimental protocols to help bridge the gap between theoretical plausibility and physical wet-lab validation. We will investigate the integration of reaction databases and computational simulation tools to enable the framework to generate not only conceptual ideas but also rigorously testable reaction pathways. Second, we plan to implement targeted control mechanisms over hypothesis complexity, allowing researchers to tailor the discovery trajectories to specific resource constraints. 

Crucially, to mitigate the dual-use risks associated with autonomous chemical discovery, developing and integrating explicit safety filters to prevent the generation of harmful or hazardous chemical pathways is a critical priority for future iterations. Finally, we intend to evaluate and fine-tune the framework's adaptability across a broader spectrum of specialized chemical sub-disciplines.

\section{Conclusion}

In this work, we introduce MAIL, a fully automated chemical hypothesis-generation framework that emphasizes dynamic literature retrieval, multi-round refinement, and adaptive prompting. By successfully navigating domain-specific constraints without manual corpus curation or human-in-the-loop interventions, MAIL autonomously generates highly novel, mechanistically sound, and structurally coherent hypotheses. Ultimately, this framework provides a robust foundation for AI-assisted chemical discovery, illustrating that iterative, literature-grounded reasoning is essential for effectively leveraging LLMs at the frontiers of scientific research.

\section*{Declaration of generative AI and AI-assisted technologies}
During the preparation of this work, the authors used large language models to assist with formatting, editing for clarity, and preparing the submission. After using this tool/service, the authors reviewed and edited the content as needed and take full responsibility for the content of the published article.

\section*{Data Availability}
The high-novelty Nature/Science challenge (HN-NS) dataset curated for this study are publicly available and can be found at https://github.com/YGanLab/MAIL. The source code for the MAIL framework will be made publicly available upon the acceptance of this manuscript.

\section*{Acknowledgments}
Financial support for this publication results from Scialog
grant \#SA-AUT-2024-022a and \#SA-AUT-2024-022b from Research Corporation for Science Advancement.

\appendix
\setcounter{table}{0}
\renewcommand{\thetable}{A.\arabic{table}}

\section{Case Study: Iterative Refinement Across Rounds}
\label{case_study_example}
We examine how \textsc{MAIL} refines an initial hypothesis over three rounds using feedback and new inspirations. We list the background knowledge, ground-truth, and generated hypotheses. Additionally, Table~\ref{tab:round_scores} shows the MIOS and MPOS scores for each round, illustrating steady alignment gains across rounds.

\subsection*{Background question} {How can we automate the optimization, intensification, and scale-up of photocatalytic reactions to enhance efficiency, reproducibility, and scalability, while reducing the need for human expertise and intervention?}

\subsection*{Background survey} {
Photocatalysis is widely used in chemical synthesis, particularly in pharmaceuticals and materials science, due to its ability to drive reactions under mild conditions using light. However, challenges persist in optimizing reaction conditions, ensuring reproducibility, and scaling up reactions from lab to industrial scale.
Traditional methods like one-factor-at-a-time (OFAT) optimization are inefficient, as they fail to account for interactions between variables like catalyst concentration, light intensity, and reaction time. While Design of Experiments (DoE) improves this by considering multiple factors, it remains time-consuming and resource-intensive, especially for complex reactions.
A significant issue in photocatalysis is inconsistent light absorption in batch reactors, which hinders scalability and leads to irreproducible results. Continuous-flow reactors offer more uniform light distribution, which improves control over mass and photon transport, making them more suitable for scale-up. However, these systems still require extensive re-optimization for different reactions due to variations in reactor design and light sources.
The complexity of photocatalytic mechanisms, often involving multiple catalytic cycles, adds another layer of difficulty in predicting optimal conditions. Traditional approaches struggle with identifying the best parameters efficiently, particularly when dealing with non-linear relationships between variables.
Recent advancements in machine learning, particularly Bayesian Optimization (BO), have shown promise in addressing these challenges by efficiently searching large experimental spaces. BO minimizes the number of experiments required to find optimal conditions, although fully integrating these methods into real-time chemical processes remains a challenge.}

\begin{table}[htbp]
\caption{Scores for each refinement round based on the two evaluation metrics.}\label{tab:round_scores}
\centering
\begin{tabular*}{\tblwidth}{@{\extracolsep{\fill}} lcc @{}}
\toprule
\textbf{Round} & \textbf{MIOS} & \textbf{MPOS} \\
\midrule
Round 1 & 1.30 & 0.00 \\
Round 2 & 3.00 & 2.00 \\
Round 3 & \textbf{4.40} & \textbf{4.00} \\
\bottomrule
\end{tabular*}
\end{table}

\subsection*{Groundtruth hypothesis} {The main hypothesis is that by integrating BO with continuous-flow reactors and real-time monitoring into an autonomous system (RoboChem), it is possible to efficiently optimize, intensify, and scale-up photocatalytic reactions with high precision and minimal human intervention. This system offers a hands-free solution to the challenges posed by manual optimization methods and batch reactors, ultimately improving reaction yield, throughput, and reproducibility.}

\subsection*{Hypothesis generated round 1} {Development of a hybrid reactor system that combines photocatalytic and microwave-assisted processes in a continuous-flow setup may synergistically improve reaction efficiency and scalability by leveraging uniform light and microwave energy distribution to enhance mass and photon transport.}

\subsection*{Hypothesis generated round 2} {Development of a hybrid continuous-flow reactor system that integrates photocatalytic and microwave-assisted processes with a Z-scheme heterojunction, such as g-C3N4@Bi /Bi2O2CO3, combined with Bayesian optimization, may enhance reaction efficiency, reproducibility, and scalability. The Z-scheme heterojunction can improve charge separation and broaden light absorption, while the synergistic use of microwave. 
Energy can enhance mass and photon transport. 
Bayesian optimization will automate parameter adjustments to navigate complex reaction spaces efficiently, reducing human intervention and ensuring consistent reaction conditions across different scales. This approach leverages advanced materials and machine learning to address the challenges of light distribution and parameter optimization in photocatalytic reactions.}

\subsection*{Hypothesis generated round 3} Development of a hybrid continuous-flow reactor system that integrates photocatalytic and microwave-assisted processes with a Z-scheme heterojunction, such as g-C3N4@Bi /Bi2O2CO3, and utilizes a Taylor Vortex Flow Reactor design to enhance mixing and mass transfer, potentially improving scalability and efficiency. This approach, combined with Bayesian optimization for real-time parameter adjustment, aims to automate and streamline the optimization of complex reaction spaces, reducing human intervention and ensuring consistent reaction conditions across different scales. The Taylor Vortex Flow Reactor's small footprint and efficient mixing capabilities address mass transfer limitations, while the integrated microwave energy and Z-scheme heterojunction improve photon and mass transport, providing a novel and robust solution for scaling up photocatalytic processes.

These results demonstrate MAIL’s capacity to progressively align an initially uninformed hypothesis toward the ground-truth target without ever having access to it. The MIOS show a steady increase, reflecting convergence in the central research idea and mechanism. Similarly, MPOS improvement indicates the incorporation of key methodological elements over successive refinement rounds. The largest performance improvement is observed between Rounds 1 and 2, when the hypothesis first incorporates "Bayesian optimization", a core methodological element of the groundtruth. Round 3 further narrows the gap by introducing "continuous-flow reactor improvements" and "real-time control", bringing the generated hypothesis into close alignment with all major groundtruth components. This pattern suggests that MAIL’s iterative feedback–inspiration loop effectively narrows conceptual and methodological gaps, moving hypotheses toward expert-designed solutions while preserving novelty.

\section{Details of the prompt design}
\label{prompts_details}
We provide detailed descriptions of the prompts used in the chemical hypothesis generation method.

\lstdefinestyle{prompt}{
  basicstyle=\ttfamily,
  breaklines=true,
  columns=fullflexible,
  showstringspaces=false,
  keywordstyle=\normalfont,
  commentstyle=\normalfont,
  frame=tb
}

\subsection{Feedback on hypothesis}
In this step, the model is tasked with acting as a highly critical domain reviewer, systematically evaluating the generated hypothesis along four research dimensions: Validness, Novelty, Significance, and Potential. The prompt provides strict 5-point scoring guidelines for each dimension, ensuring the evaluation is detailed, consistent, and actionable for guiding the next refinement iteration.
\lstset{style=prompt}
\begin{lstlisting}
You are known as a diligent and super harsh reviewer in Chemistry and Materials Science who will spend much time to find flaws when reviewing and therefore usually gives a relatively much lower score than other reviewers. But when you meet with a hypothesis you truly appreciate, you don't mind giving it good scores. Given a not-yet-peer-reviewed research hypothesis in the Chemistry and Materials Science domain, try to evaluate the research hypothesis from four research aspects and give a score according to the evaluation guidelines provided below. All four aspects should be evaluated on a 5-point scale.

Aspect 1: Validness.  
5 points: The hypothesis is a logical next step from current research, strongly supported by theory, perhaps with some indirect experimental evidence or highly predictive computational results. The experimental verification seems straightforward with a high probability of confirming the hypothesis.  
4 points: Here, the hypothesis is well-rooted in existing theory with some preliminary data or computational models supporting it. It extends known science into new but logically consistent areas, where experiments are feasible with current technology, and there's a reasonable expectation of positive results.  
3 points: This hypothesis is within the realm of theoretical possibility but stretches the boundaries of what's known. It might combine existing knowledge in very novel ways or predict outcomes for which there's no direct evidence yet. There's a conceptual framework for testing, but success is uncertain.  
2 points: While the hypothesis might be grounded in some theoretical aspects, it significantly deviates from current understanding or requires conditions or materials that are currently impossible or highly improbable to achieve or synthesize.  
1 point: The hypothesis proposes concepts or outcomes that are not only unsupported by current theory but also contradict well-established principles or data. There's no clear path to experimental testing due to fundamental theoretical or practical barriers.  

Aspect 2: Novelty.  
5 points: This level of novelty could fundamentally alter our understanding of Chemistry and Material Science or create entirely new fields. It often involves predictions or discoveries that, if proven, would require a significant overhaul of existing Chemistry and Materials Science theories.  
4 points: The hypothesis significantly departs from established norms, potentially redefining how certain Chemistry and Material Science phenomena are understood or applied. It might involve entirely new materials or theoretical frameworks;  
3 points: This level involves a hypothesis that could potentially lead to new insights or applications. It might challenge minor aspects of current theories or introduce new methodologies or materials;  
2 points: The hypothesis introduces a new angle or method within an established framework. It might involve known compounds or reactions, but in contexts or combinations not previously explored.  
1 point: The hypothesis involves minor tweaks or applications of well-known principles or techniques. It might slightly extend existing knowledge, but it doesn't introduce fundamentally new concepts.  

Aspect 3: Significance.  
5 points: This hypothesis could fundamentally change one or more branches of Chemistry and Materials Science. It might introduce entirely new principles, theories, or methodologies that redefine the boundaries of Chemistry and Material Science;  
4 points: This hypothesis challenges current understanding or introduces a concept that could lead to substantial changes in how a particular area of Chemistry and Materials Science is viewed or applied. It might lead to new technologies or significant theoretical advancement;  
3 points: this hypothesis proposes something new or an innovative approach that could lead to noticeable advancements in a specific area of Chemistry and Materials Science. It might open new avenues for research or application, but it doesn't revolutionize the field.;  
2 points: This hypothesis might offer a small variation or incremental improvement on existing knowledge. It could potentially refine a known concept, but doesn't significantly alter the field.;  
1 point: The hypothesis addresses a very narrow or already well-established aspect of Chemistry. It might confirm what is already known without adding much new insight.  

Aspect 4: Potential.  
5 points: The hypothesis, while potentially intriguing now, holds the promise of being revolutionary with the addition of a key methodological component. This could introduce entirely new concepts or fields, fundamentally changing our understanding or capabilities in Chemistry and Materials Science.  
4 points: The hypothesis, though promising, could be transformative with the right methodological enhancement. This enhancement might lead to groundbreaking discoveries or applications, significantly advancing the field.  
3 points: The hypothesis, while interesting in its current form, could be significantly elevated with the right methodological addition. This might lead to new insights or applications that go beyond the initial scope.  
2 points: The hypothesis currently offers some value but has the potential for more substantial contributions if enhanced with a new methodological approach. This could lead to incremental advancements in understanding or application.  
1 point: The hypothesis, as it stands, might be straightforward or well-trodden. Even with methodological enhancements, it's unlikely to significantly expand current knowledge or applications beyond minor improvements.

The hypothesis is:

Please give a response to the initial question on scoring the hypothesis from four aspects. Remember that you are a diligent and harsh reviewer. (response format: 'Concise reason for validness score:  
Validness score:  
Concise reason for novelty score:  
Novelty score:  
Concise reason for significance score:  
Significance score:  
Concise reason for potential score:  
Potential score:').
The hypothesis is: "{hypo}"

\end{lstlisting}

\subsection{Inspiration paper selection}
In this stage, the model selects the most relevant inspiration papers to guide hypothesis refinement in subsequent rounds. The prompt explicitly instructs the model to choose papers that address weaknesses identified in feedback and contribute mechanisms, methods, or concepts that improve the novelty, specificity, or plausibility of the hypothesis.
\lstset{style=prompt}
\begin{lstlisting}
You are assisting in a scientific hypothesis refinement task.

A research hypothesis was generated from the background context and prior literature. It has received critical feedback, and your job is to identify new papers that can help **revise or strengthen** it.

---

## TASK CONTEXT

**Research Question:**  
{bg_question}

**Background Summary:**  
{bg_survey}

**Current Hypothesis:**  
{hypo}

**Feedback on Hypothesis:**  
{feedback}

**Candidate Inspiration Papers:**  
{inspiration_texts}

**Accumulated Memory Summary**
{memory_summary}
---

## YOUR TASK

Select the **two best inspiration papers** that could help address the issues identified in the feedback or improve the current hypothesis. Each selected paper should:

- Introduce a **mechanism**, method, or concept** that directly relates to fixing one or more weaknesses in the hypothesis.
- Offer **novel direction**, clarify a **missing mechanism**, or support a more **specific or plausible design**.

---

## OUTPUT FORMAT (STRICT)

Respond in the format below with exactly two selections:
(response format: 'Title: 
Reason: 
Title: 
Reason: 
')
\end{lstlisting}

\subsection{Hypothesis refinement}
In the hypothesis refinement stage, the model incorporates structured feedback from the previous round together with newly selected inspiration papers to produce an improved version of the hypothesis. The prompt explicitly instructs the model to address weaknesses identified in prior evaluations, integrate at least one relevant mechanism or method from the new inspirations, and preserve strong elements from earlier versions.
\lstset{style=prompt}
\begin{lstlisting}
You are a scientific reasoning assistant specializing in hypothesis refinement.

Your task is to update a previously proposed research hypothesis using new feedback and inspirations. If the original hypothesis already meets high standards of novelty, specificity, validity, and significance (per feedback), you may return it unchanged. Otherwise, improve it by addressing weaknesses and integrating new inspirational materials.

---

### INPUTS

**1. Background Research Question**  
{bg_question}

**2. Background Survey**  
{bg_survey}

**3. Hypothesis from Round {round_num - 1}**  
{hypo}

**4. Feedback on Hypothesis (Round {round_num})**  
The hypothesis was evaluated based on: **Specificity**, **Novelty**, **Validity**, and **Significance**. Below is the structured feedback:
{feedback}

**5. New Inspiration Articles (Round {round_num})**  
This paper was selected specifically to help improve the current hypothesis based on feedback.
{inspiration_texts}

**6. Accumulated Memory Summary**
{memory_summary}
---

### YOUR TASK

Based on the inputs above, generate **one revised research hypothesis** that:
- Addresses the core research question
- Fixes issues identified in the feedback (e.g., vagueness, lack of novelty, or mechanism)
- Integrates at least one idea, mechanism, or method from the new inspiration articles
- Uses clear, mechanistic reasoning to describe **how** and **why** the hypothesis works
- Is concise (2-5 sentences) and testable with a plausible experimental design

Avoid simply rephrasing the original hypothesis. If reusing content, justify it by alignment with positive feedback. If rewriting, make the improvements traceable to the feedback and inspiration.

---

### OUTPUT FORMAT (STRICT)

Respond with a **single, numbered hypothesis** like this:

1. <Your revised hypothesis here>
\end{lstlisting}

\subsection{Hypothesis selection}
In the final selection stage, multiple refined hypotheses from different iterations are compared to identify the most promising candidate.
The model evaluates each hypothesis using accumulated memory from prior rounds and applies structured criteria (novelty, specificity, validity, and significance) to guide its decision.
The prompt instructs the model to weigh trade-offs across these criteria and choose the hypothesis with the strongest overall scientific merit.
\lstset{style=prompt}
\begin{lstlisting}
You are tasked with selecting the research hypothesis that best aligns with the research question and demonstrates the highest potential for scientific innovation and impact. Use the refined criteria below to guide your selection:

1. **Alignment with Research Question**: Assess how directly and effectively the hypothesis addresses the core elements of the research question, ensuring it is relevant and coherent.
2. **Empirical Support**: Evaluate the hypothesis's foundation on empirical evidence, emphasizing robust citation of relevant studies and potential for validation through practical experiments.
3. **Feasibility**: Examine the clarity and practicality of the proposed methods, considering the availability of resources and techniques, as well as the likelihood of successful implementation.
4. **Innovative Potential**: Consider the novelty and potential for breakthroughs, focusing on unique methodologies or combinations that could lead to discoveries.
5. **Impact Potential**: Analyze the hypothesis's potential to deliver transformative insights and advancements, with attention to its broader implications and applications.

** Input: **
- **Research Question: {bg_question}
- **Background Survey: {bg_survey}
- **Input Hypotheses: {hypos}
- **Accumulated Memory Summary: {memory_summary}

Evaluate each hypothesis based on these criteria, prioritizing those that utilize insights from the background survey, propose innovative solutions, and demonstrate strong empirical support, clear feasibility, and significant impact potential. Select the hypothesis that aligns most closely with these criteria and exhibits the highest potential for scientific advancement.

** Selection Strategy: **
- Prioritize hypotheses with compelling empirical evidence and a clear pathway to validation.
- Choose hypotheses offering innovative and practical methods that effectively incorporate insights from the background survey.
- Ensure the hypothesis is directly relevant to the research question and demonstrates the highest potential for scientific impact and innovation.

Evaluate based on these criteria, focusing on selecting the hypothesis with the highest score for success and alignment with the research question.

** Output Format: **
1. <Selected Hypothesis>
\end{lstlisting}

\subsection{Automatic evaluation}
To assess the quality of generated hypotheses without relying solely on human evaluation, we employ an automatic evaluation pipeline using GPT-4o. The model is instructed to score each hypothesis according to multiple structured criteria while also providing concise, evidence-based reasoning.
Matched Insight Overlap Score (MIOS) measures the conceptual similarity and scientific relevance of a generated hypothesis to the groundtruth, and Matched Point Overlap Score (MPOS) measures the extent to which the generated hypothesis covers the methodological key points present in the groundtruth.

\lstset{style=prompt}
\begin{lstlisting}
You are an expert scientific evaluator. Rate how well the generated hypothesis matches the main idea of the groundtruth hypothesis in chemistry.

Focus only on conceptual alignment of the primary claim and mechanism (the central research idea, goal, and high-level mechanistic rationale),
not on writing quality, length, citations, or minor experimental details. Paraphrases are acceptable. Penalize contradictions or shifts in scope.

Scoring rubric (MIOS, 1.0-5.0; decimals allowed):
5.0  = Nearly the same main idea: same objective and high-level mechanism/logic; only minor phrasing/detail differences.
4.x  = Strongly aligned main idea with small deviations in mechanism or scope.
3.x  = Partially aligned: overlaps on objective or mechanism but not both; noticeable gaps or additions.
2.x  = Weak alignment: only tangentially related theme; major differences in objective and mechanism.
1.x  = Unrelated or contradicts the groundtruth's main idea.

Generated Hypothesis:
{generated}

Groundtruth Hypothesis:
{groundtruth}

Please evaluate the proposed hypothesis based on the groundtruth hypothesis, and give a score. (response format: 'Reason: 
Matched score: <score>
')
\end{lstlisting}

\lstset{style=prompt}
\begin{lstlisting}
You are helping to evaluate the quality of a proposed research hypothesis in Chemistry by a PhD student. The groundtruth
hypothesis will also be provided to compare. Here, we mainly focus on whether the proposed hypothesis has covered the key
points in terms of the methodology in the groundtruth hypothesis. You will also be given a summary of the key points in
the methodology of the groundtruth hypothesis for reference. Please note that for the proposed hypothesis to cover one key
point, it is not necessary to explicitly mention the name of the key point, but it might also integrate the key point implicitly
in the proposed method. The evaluation criterion is called 'Matched score', which is on a 6-point Likert scale (from 5 to 0).
Particularly, 5 points mean that the proposed hypothesis (1) covers all the key points and leverages them similarly as in the
methodology of the groundtruth hypothesis, and (2) does not contain any extra key point that has apparent flaws; 4 points mean
that the proposed hypothesis (1) covers all the key points (or at least three key points) and leverages them similarly as in the
methodology of the groundtruth hypothesis, (2) but also with extra key points that have apparent flaws; 3 points mean that the
proposed hypothesis (1) covers at least two key points and leverages them similarly as in the methodology of the groundtruth
hypothesis, (2) but does not cover all key points in groundtruth hypothesis, (3) might or might not contain extra key points
points; 2 points mean that the proposed hypothesis (1) covers at least one key point in the methodology of the groundtruth
hypothesis, and leverage it similarly as in the methodology of groundtruth hypothesis, (2) but does not cover all key points in
the groundtruth hypothesis, and (3) might or might not contain extra key points; 1 point means that the proposed hypothesis (1)
covers at least one key point in the methodology of the groundtruth hypothesis, (2) but is used differently in the methodology
of groundtruth hypothesis, and (3) might or might not contain extra key points; 0 point means that the proposed hypothesis
does not cover any key point in the methodology of the groundtruth hypothesis at all. Please note that the total number of keys
points in the groundtruth hypothesis might be less than three, so that multiple points can be given. E.g., there's only one key
point in the groundtruth hypothesis, and the proposed hypothesis covers the one key point, it's possible to give 2 points, 4
points, and 5 points. In this case, we should choose a score from 4 points to 5 points, depending on the existence and quality
of extra key points. 'Leveraging a key point similarly as in the methodology of the groundtruth hypothesis' means that in the
proposed hypothesis, the same (or very related) concept (key point) is used similarly with a similar goal compared to
the groundtruth hypothesis (not necessarily for the proposed hypothesis to be the same as the groundtruth hypothesis
to be classified as 'similar'). When judging whether an extra key point has apparent flaws, you should use your own knowledge
to judge, but rather than to rely on the number of extra key points to judge.
Please evaluate the proposed hypothesis based on the groundtruth hypothesis.
---

**Groundtruth Hypothesis**:
{groundtruth}

**Proposed Hypothesis**:
{generated}

**The key points in the groundtruth hypothesis are:**
{keypoints}
---

Please evaluate the proposed hypothesis based on the groundtruth hypothesis, and give a score. (response format: 'Reason: 
Matched score: <score>
')
\end{lstlisting}

\bibliographystyle{unsrtnat}
\bibliography{custom}

@article{wei2022chain,
  title={Chain-of-thought prompting elicits reasoning in large language models},
  author={Wei, Jason and Wang, Xuezhi and Schuurmans, Dale and Bosma, Maarten and Xia, Fei and Chi, Ed and Le, Quoc V and Zhou, Denny and others},
  journal={Advances in neural information processing systems},
  volume={35},
  pages={24824--24837},
  year={2022},
  url = {https://dl.acm.org/doi/10.5555/3600270.3602070}
}

@article{akujuobi2024link,
  title = {Link prediction for hypothesis generation: an active curriculum learning infused temporal graph-based approach},
  author = {Akujuobi, Uchenna and Kumari, Priyadarshini and Choi, Jihun and Badreddine, Samy and Maruyama, Kana and Palaniappan, Sucheendra K. and Besold, Tarek R.},
  journal = {Artificial Intelligence Review},
  volume = {57},
  number = {9},
  pages = {244},
  year = {2024},
  publisher = {Springer},
  doi = {10.1007/s10462-024-10885-1}
}

@techreport{white2019publications,
  author = {White, Karen},
  title = {Publications Output: {U.S.} Trends and International Comparisons},
  institution = {National Science Board},
  number = {NSB-2020-6},
  type = {Science \& Engineering Indicators 2020},
  year = {2019},
  month = dec,
  address = {Alexandria, VA},
  url = {https://ncses.nsf.gov/pubs/nsb20206}
}

@article{zhao2024survey,
  title = {A Survey on Evaluation of Large Language Models},
  author = {Chang, Yupeng and Wang, Xu and Wang, Jindong and Wu, Yuan and Yang, Linyi and Zhu, Kaijie and Chen, Hao and Yi, Xiaoyuan and Wang, Cunxiang and Wang, Yidong and Ye, Wei and Zhang, Yue and Chang, Yi and Yu, Philip S. and Yang, Qiang and Xie, Xing},
  year = {2024},
  journal = {{ACM} Transactions on Intelligent Systems and Technology},
  volume = {15},
  number = {3},
  pages = {39:1--39:45},
  doi = {10.1145/3641289},
  publisher = {Association for Computing Machinery}
}

@article{qi2023large,
  title   = {{L}arge language models are zero-shot hypothesis proposers},
  author  = {Qi, Biqing and Zhang, Kaiyan and Li, Haoxiang and Tian, Kai and Zeng, Sihang and Chen, Zhang-Ren and Zhou, Bowen},
  journal = {arXiv preprint arXiv:2311.05965},
  year    = {2023},
  url     = {https://arxiv.org/abs/2311.05965}
}

@inproceedings{kumar-etal-2025-large,
  title     = {Can Large Language Models Unlock Novel Scientific Research Ideas?},
  author    = {Kumar, Sandeep and Ghosal, Tirthankar and Goyal, Vinayak and Ekbal, Asif},
  booktitle = {Proceedings of the 2025 Conference on Empirical Methods in Natural Language Processing (EMNLP)},
  pages     = {33551--33575},
  month     = nov,
  year      = {2025},
  address   = {Suzhou, China},
  publisher = {Association for Computational Linguistics},
  doi       = {10.18653/v1/2025.emnlp-main.1704},
  url       = {https://aclanthology.org/2025.emnlp-main.1704}
}

@article{yang2024moose,
  title   = {{MOOSE-Chem}: {L}arge language models for rediscovering unseen chemistry scientific hypotheses},
  author  = {Yang, Zonglin and Liu, Wanhao and Gao, Ben and Xie, Tong and Li, Yuqiang and Ouyang, Wanli and Poria, Soujanya and Cambria, Erik and Zhou, Dongzhan},
  journal = {arXiv preprint arXiv:2410.07076},
  year    = {2024},
  url     = {https://arxiv.org/abs/2410.07076}
}

@article{mirza2025framework,
  title   = {A framework for evaluating the chemical knowledge and reasoning abilities of large language models against the expertise of chemists},
  author  = {Mirza, Adrian and Alampara, Nawaf and Kunchapu, Sreekanth and R{\'\i}os-Garc{\'\i}a, Marti{\~n}o and Emoekabu, Benedict and Krishnan, Aswanth and Gupta, Tanya and Schilling-Wilhelmi, Mara and Okereke, Macjonathan and Aneesh, Anagha},
  journal = {Nature Chemistry},
  pages   = {1--8},
  year    = {2025},
  publisher = {Nature Publishing Group {UK} London},
  doi     = {10.1038/s41557-025-01815-x}
}

@inproceedings{yang2023large,
  title     = {Large Language Models for Automated Open-domain Scientific Hypotheses Discovery},
  author    = {Yang, Zonglin and Du, Xinya and Li, Junxian and Zheng, Jie and Poria, Soujanya and Cambria, Erik},
  booktitle = {Findings of the Association for Computational Linguistics ({ACL} 2024)},
  month     = aug,
  year      = {2024},
  address   = {Bangkok, Thailand},
  publisher = {Association for Computational Linguistics},
  doi       = {10.18653/v1/2024.findings-acl.804},
  url       = {https://aclanthology.org/2024.findings-acl.804},
  pages     = {13545--13565}
}

@article{swanson1986fish,
  title   = {Fish oil, {R}aynaud's syndrome, and undiscovered public knowledge},
  author  = {Swanson, Don R.},
  journal = {Perspectives in Biology and Medicine},
  volume  = {30},
  number  = {1},
  pages   = {7--18},
  year    = {1986},
  doi     = {10.1353/pbm.1986.0087}
}

@inproceedings{spangler2014automated,
  title     = {Automated hypothesis generation based on mining scientific literature},
  author    = {Spangler, Scott and Wilkins, Angela D. and Bachman, Benjamin J. and others},
  booktitle = {Proceedings of the 20th {ACM} {SIGKDD} International Conference on Knowledge Discovery and Data Mining ({KDD})},
  pages     = {1877--1886},
  year      = {2014},
  doi       = {10.1145/2623330.2623667}
}

@article{belford2018stability,
  title   = {Stability of topic modeling via matrix factorization},
  author  = {Belford, Mark and Mac Namee, Brian and Greene, Derek},
  journal = {Expert Systems with Applications},
  volume  = {91},
  pages   = {159--169},
  year    = {2018},
  doi     = {10.1016/j.eswa.2017.08.047}
}

@article{radensky2024scideator,
  title   = {{S}ci{D}eator: Human-{LLM} scientific idea generation grounded in research-paper facet recombination},
  author  = {Radensky, Marissa and Shahid, Simra and Fok, Raymond and Siangliulue, Pao and Hope, Tom and Weld, Daniel S.},
  journal = {arXiv preprint arXiv:2409.14634},
  year    = {2024},
  url     = {https://arxiv.org/abs/2409.14634}
}

@inproceedings{wang2024scimon,
  title     = {{S}ci{MON}: Scientific Inspiration Machines Optimized for Novelty},
  author    = {Wang, Qingyun and Downey, Doug and Ji, Heng and Hope, Tom},
  booktitle = {Proceedings of the 62nd Annual Meeting of the Association for Computational Linguistics ({ACL} 2024)},
  pages     = {279--299},
  year      = {2024},
  url       = {https://aclanthology.org/2024.acl-long.17}
}

@article{sprueill2402chemreasoner,
  title   = {{C}hem{R}easoner: Heuristic Search over a Large Language Model's Knowledge Space Using Quantum-Chemical Feedback},
  author  = {Sprueill, H. W. and Edwards, C. and Agarwal, K. and others},
  journal = {arXiv preprint arXiv:2402.10980},
  year    = {2024},
  url     = {https://arxiv.org/abs/2402.10980}
}

@article{wang2024chain,
  title   = {Chain-of-thought Reasoning without Prompting},
  author  = {Wang, Xuezhi and Zhou, Denny},
  journal = {Advances in Neural Information Processing Systems ({N}eur{IPS} 2024)},
  volume  = {37},
  pages   = {66383--66409},
  year    = {2024},
  url = {http://dl.acm.org/doi/10.5555/3737916.3740039}
}

@inproceedings{yao2023tree,
  title     = {{T}ree of {T}houghts: Deliberate Problem Solving with Large Language Models},
  author    = {Yao, Shunyu and Yu, Dian and Zhao, Jeffrey and others},
  booktitle = {Advances in Neural Information Processing Systems ({N}eur{IPS} 2023)},
  year      = {2023},
  pages     = {11809–11822},
  url       = {https://openreview.net/forum?id=5Xc1ecxO1h}
}

@inproceedings{shinn2023reflexion,
author = {Shinn, Noah and Cassano, Federico and Gopinath, Ashwin and Narasimhan, Karthik and Yao, Shunyu},
title = {Reflexion: language agents with verbal reinforcement learning},
year = {2023},
publisher = {Curran Associates Inc.},
address = {Red Hook, NY, USA},
booktitle = {Proceedings of the 37th International Conference on Neural Information Processing Systems},
articleno = {377},
pages     = {8634–8652},
numpages = {19},
location = {New Orleans, LA, USA},
series = {NIPS '23}
}

@inproceedings{lewis2021retrievalaugmentedgenerationknowledgeintensivenlp,
  title     = {Retrieval-Augmented Generation for Knowledge-Intensive {NLP} Tasks},
  author    = {Lewis, Patrick and Perez, Ethan and Piktus, Aleksandra and others},
  booktitle = {Advances in Neural Information Processing Systems ({N}eur{IPS} 2020)},
  volume    = {33},
  pages     = {9459--9475},
  year      = {2020},
  url       = {https://proceedings.neurips.cc/paper/2020/hash/6b493230205f780e1bc26945df7481e5-Abstract.html}
}

@inproceedings{guu2020realmretrievalaugmentedlanguagemodel,
  title     = {{REALM}: Retrieval-Augmented Language Model Pre-Training},
  author    = {Guu, Kelvin and Lee, Kenton and Tung, Zora and Pasupat, Panupong and Chang, Ming-Wei},
  booktitle = {Proceedings of the 37th International Conference on Machine Learning ({ICML} 2020)},
  pages     = {3929--3938},
  year      = {2020},
  url       = {https://proceedings.mlr.press/v119/guu20a.html}
}

@article{LU20111150,
  title   = {Link prediction in complex networks: {A} survey},
  author  = {L{\"u}, Linyuan and Zhou, Tao},
  journal = {Physica A: Statistical Mechanics and its Applications},
  volume  = {390},
  number  = {6},
  pages   = {1150--1170},
  year    = {2011},
  doi     = {10.1016/j.physa.2010.11.027}
}

@article{xiong2024improvingscientifichypothesisgeneration,
  title   = {Improving Scientific Hypothesis Generation with Knowledge Grounded Large Language Models},
  author  = {Xiong, Guangzhi and Xie, Eric and Shariatmadari, Amir Hassan and Guo, Sikun and Bekiranov, Stefan and Zhang, Aidong},
  journal = {arXiv preprint arXiv:2411.02382},
  year    = {2024},
  url     = {https://arxiv.org/abs/2411.02382}
}

@article{alkan2025surveyhypothesisgenerationscientific,
  title   = {A Survey on Hypothesis Generation for Scientific Discovery in the Era of Large Language Models},
  author  = {Alkan, Atilla Kaan and Sourav, Shashwat and Jablonska, Maja and Astarita, Simone and Chakrabarty, Rishabh and Garuda, Nikhil and Khetarpal, Pranav and Pi{\'o}ro, Maciej and Tanoglidis, Dimitrios and Iyer, Kartheik G. and Polimera, Mugdha S. and Smith, Michael J. and Ghosal, Tirthankar and Huertas-Company, Marc and Kruk, Sandor and Schawinski, Kevin and Ciuc{\u a}, Ioana},
  journal = {arXiv preprint arXiv:2504.05496},
  year    = {2025},
  url     = {https://arxiv.org/abs/2504.05496}
}

@article{10.1093/bioinformatics/btae353,
  title     = {{KRAGEN}: A Knowledge Graph-Enhanced {RAG} Framework for Biomedical Problem Solving Using Large Language Models},
  author    = {Matsumoto, Nicholas and Moran, Jay and Choi, Hyunjun and Hernandez, Miguel E. and Venkatesan, Mythreye and Wang, Paul and Moore, Jason H.},
  journal   = {Bioinformatics},
  volume    = {40},
  number    = {6},
  pages     = {btae353},
  year      = {2024},
  month     = jun,
  doi       = {10.1093/bioinformatics/btae353}
}

@article{delile2024graphbasedretrievercaptureslong,
  title   = {Graph-Based Retriever Captures the Long Tail of Biomedical Knowledge},
  author  = {Delile, Julien and Mukherjee, Srayanta and Van Pamel, Anton and Zhukov, Leonid},
  journal = {arXiv preprint arXiv:2402.12352},
  year    = {2024},
  url     = {https://arxiv.org/abs/2402.12352}
}

@article{huang2024multimodal,
  title   = {Multimodal Task Vectors Enable Many-Shot Multimodal In-Context Learning},
  author  = {Huang, Brandon and Mitra, Chancharik and Arbelle, Assaf and Karlinsky, Leonid and Darrell, Trevor and Herzig, Roei},
  journal = {Advances in Neural Information Processing Systems ({N}eur{IPS} 2024)},
  volume  = {37},
  pages   = {22124--22153},
  year    = {2024},
  url={https://proceedings.neurips.cc/paper_files/paper/2024/file/27571b74d6cd650b8eb6cf1837953ae8-Paper-Conference.pdf}
}

@inproceedings{liu2024context,
  title     = {In-Context Learning for Zero-Shot Medical Report Generation},
  author    = {Liu, Rui and Li, Mingjie and Zhao, Shen and Chen, Ling and Chang, Xiaojun and Yao, Lina},
  booktitle = {Proceedings of the 32nd {ACM} International Conference on Multimedia},
  pages     = {8721--8730},
  year      = {2024},
  doi       = {10.1145/3664647.3681436}
}

@inproceedings{hu2025nova,
    title = "{NOVA}: An Iterative Planning Framework for Enhancing Scientific Innovation with Large Language Models",
    author = "Hu, Xiang  and
      Fu, Hongyu  and
      Wang, Jinge  and
      Wang, Yifeng  and
      Li, Zhikun  and
      Xu, Renjun  and
      Lu, Yu  and
      Jin, Yaochu  and
      Pan, Lili  and
      Lan, Zhenzhong",
    editor = "Che, Wanxiang  and
      Nabende, Joyce  and
      Shutova, Ekaterina  and
      Pilehvar, Mohammad Taher",
    booktitle = "Findings of the Association for Computational Linguistics: ACL 2025",
    month = jul,
    year = "2025",
    address = "Vienna, Austria",
    publisher = "Association for Computational Linguistics",
    url = "https://aclanthology.org/2025.findings-acl.1099/",
    doi = "10.18653/v1/2025.findings-acl.1099",
    pages = "21330--21359",
    ISBN = "979-8-89176-256-5",
}

@article{liu2025researchbench,
  title   = {{ResearchBench}: Benchmarking {LLM}s in Scientific Discovery via Inspiration-Based Task Decomposition},
  author  = {Liu, Yujie and Yang, Zonglin and Xie, Tong and Ni, Jinjie and Gao, Ben and Li, Yuqiang and Tang, Shixiang and Ouyang, Wanli and Cambria, Erik and Zhou, Dongzhan},
  journal = {arXiv preprint arXiv:2503.21248},
  year    = {2025},
  url     = {https://arxiv.org/abs/2503.21248}
}

@article{noh2024retrieval,
  title={Retrieval-Retro: Retrieval-based Inorganic Retrosynthesis with Expert Knowledge},
  author={Noh, Heewoong and Lee, Namkyeong and Na, Gyoung S and Park, Chanyoung},
  journal={Advances in Neural Information Processing Systems},
  volume={37},
  pages={25375--25400},
  year={2024},
  url={https://proceedings.neurips.cc/paper_files/paper/2024/file/2cfa9b0d9be8a5c01cf3eb7f21b4f2b8-Paper-Conference.pdf}
}

@article{OTYEPKA2025102981,
title = {Advancing materials discovery through artificial intelligence},
journal = {Applied Materials Today},
volume = {47},
pages = {102981},
year = {2025},
issn = {2352-9407},
doi = {https://doi.org/10.1016/j.apmt.2025.102981},
url = {https://www.sciencedirect.com/science/article/pii/S2352940725003981},
author = {Martin Otyepka and Martin Pykal and Michal Otyepka},
}

@article{zhang2025exploring,
  title={Exploring the role of large language models in the scientific method: from hypothesis to discovery},
  author={Zhang, Yanbo and Khan, Sumeer A and Mahmud, Adnan and Yang, Huck and Lavin, Alexander and Levin, Michael and Frey, Jeremy and Dunnmon, Jared and Evans, James and Bundy, Alan and others},
  journal={npj Artificial Intelligence},
  volume={1},
  number={1},
  pages={14},
  year={2025},
  url = {https://www.nature.com/articles/s44387-025-00019-5}
}

@inproceedings{bai-etal-2024-advancing,
  title={Advancing abductive reasoning in knowledge graphs through complex logical hypothesis generation},
  author={Bai, Jiaxin and Wang, Yicheng and Zheng, Tianshi and Guo, Yue and Liu, Xin and Song, Yangqiu},
  booktitle={Proceedings of the 62nd Annual Meeting of the Association for Computational Linguistics (Volume 1: Long Papers)},
  pages={1312--1329},
  year={2024},
  publisher = {Association for Computational Linguistics},
  url = {https://aclanthology.org/2024.acl-long.72/},
}

@inproceedings{zheng-etal-2025-automation,
    title = "From Automation to Autonomy: A Survey on Large Language Models in Scientific Discovery",
    author = "Zheng, Tianshi  and
      Deng, Zheye  and
      Tsang, Hong Ting  and
      Wang, Weiqi  and
      Bai, Jiaxin  and
      Wang, Zihao  and
      Song, Yangqiu",
    editor = "Christodoulopoulos, Christos  and
      Chakraborty, Tanmoy  and
      Rose, Carolyn  and
      Peng, Violet",
    booktitle = "Proceedings of the 2025 Conference on Empirical Methods in Natural Language Processing",
    month = nov,
    year = "2025",
    address = "Suzhou, China",
    publisher = "Association for Computational Linguistics",
    url = "https://aclanthology.org/2025.emnlp-main.895/",
    doi = "10.18653/v1/2025.emnlp-main.895",
    pages = "17733--17750",
    ISBN = "979-8-89176-332-6"
}

@inproceedings{liu-etal-2025-literature,
    title = "Literature Meets Data: A Synergistic Approach to Hypothesis Generation",
    author = "Liu, Haokun  and
      Zhou, Yangqiaoyu  and
      Li, Mingxuan  and
      Yuan, Chenfei  and
      Tan, Chenhao",
    editor = "Che, Wanxiang  and
      Nabende, Joyce  and
      Shutova, Ekaterina  and
      Pilehvar, Mohammad Taher",
    booktitle = "Proceedings of the 63rd Annual Meeting of the Association for Computational Linguistics (Volume 1: Long Papers)",
    month = jul,
    year = "2025",
    address = "Vienna, Austria",
    publisher = "Association for Computational Linguistics",
    url = "https://aclanthology.org/2025.acl-long.12/",
    doi = "10.18653/v1/2025.acl-long.12",
    pages = "245--281",
    ISBN = "979-8-89176-251-0",
}

@article{BORREGO2025113280,
title = {Research hypothesis generation over scientific knowledge graphs},
journal = {Knowledge-Based Systems},
volume = {315},
pages = {113280},
year = {2025},
issn = {0950-7051},
doi = {https://doi.org/10.1016/j.knosys.2025.113280},
url = {https://www.sciencedirect.com/science/article/pii/S0950705125003272},
author = {Agustín Borrego and Danilo Dessì and Daniel Ayala and Inma Hernández and Francesco Osborne and Diego {Reforgiato Recupero} and Davide Buscaldi and David Ruiz and Enrico Motta},
}

@article{che2026select,
  title={Select Prompting with Chain-of-Thought paired with Large Language Models},
  author={Che, Xun and Wu, Wenjia and Chen, Yadang and Jiang, Luanjuan and Li, Qianmu},
  journal={Expert Systems with Applications},
  pages={131511},
  year={2026},
  publisher={Elsevier}
}

@article{oyelade2025smar+,
  title={SMAR+ NIE IdeaGen: A knowledge graph based node importance estimation with analogical reasoning on large language model for idea generation},
  author={Oyelade, Olaide N and Wang, Hui and Rafferty, Karen},
  journal={Expert Systems with Applications},
  volume={279},
  pages={127455},
  year={2025},
  publisher={Elsevier}
}

@article{van2025new,
  title={A New Benchmark Dataset and Mixture-of-Experts Language Models for Adversarial Natural Language Inference in Vietnamese},
  author={Van Huynh, Tin and Van Nguyen, Kiet and Nguyen, Ngan Luu-Thuy},
  journal={Expert Systems with Applications},
  pages={130109},
  year={2025},
  publisher={Elsevier}
}

@article{dubey2024llama,
  title={The llama 3 herd of models},
  author={Grattafiori, Aaron and Dubey, Abhimanyu and Jauhri, Abhinav and Pandey, Abhinav and Kadian, Abhishek and Al-Dahle, Ahmad and Letman, Aiesha and Mathur, Akhil and Schelten, Alan and Vaughan, Alex and others},
  journal={arXiv preprint arXiv:2407.21783},
  year={2024}
}

@misc{qwen2.5,
    title = {Qwen2.5: A Party of Foundation Models},
    url = {https://qwenlm.github.io/blog/qwen2.5/},
    author = {Qwen Team},
    month = {September},
    year = {2024}
}

@inproceedings{you2018gcpn,
  title     = {Graph Convolutional Policy Network for Goal-Directed Molecular Graph Generation},
  author    = {You, Jiaxuan and Liu, Bowen and Ying, Zhitao and
               Pande, Vijay and Leskovec, Jure},
  booktitle = {Advances in Neural Information Processing Systems},
  volume    = {31},
  pages     = {6410--6421},
  year      = {2018},
  url       = {https://proceedings.neurips.cc/paper/2018/hash/d60678e8f2ba9c540798ebbde31177e8-Abstract.html}
}

@inproceedings{bengio2021gflownet,
  title     = {Flow Network Based Generative Models for Non-Iterative Diverse Candidate Generation},
  author    = {Bengio, Emmanuel and Jain, Moksh and Korablyov, Maksym and
               Precup, Doina and Bengio, Yoshua},
  booktitle = {Advances in Neural Information Processing Systems},
  volume    = {34},
  pages     = {27381--27394},
  year      = {2021},
  url       = {https://proceedings.neurips.cc/paper/2021/hash/e614f646836aaed9f89ce58e837e2310-Abstract.html}
}

@article{bengio2023gflownet,
  title   = {{GFlowNet} Foundations},
  author  = {Bengio, Yoshua and Lahlou, Salem and Deleu, Tristan and
             Hu, Edward J. and Tiwari, Mo and Bengio, Emmanuel},
  journal = {Journal of Machine Learning Research},
  volume  = {24},
  number  = {210},
  pages   = {1--55},
  year    = {2023},
  url     = {https://jmlr.org/papers/v24/22-0364.html}
}

@inproceedings{kumbhar-etal-2025-hypothesis,
  title     = {Hypothesis Generation for Materials Discovery and Design Using Goal-Driven and Constraint-Guided {LLM} Agents},
  author    = {Kumbhar, Shrinidhi and Mishra, Venkatesh and
               Coutinho, Kevin and Handa, Divij and
               Iquebal, Ashif and Baral, Chitta},
  booktitle = {Findings of the Association for Computational Linguistics: NAACL 2025},
  pages     = {7539--7570},
  year      = {2025},
  doi       = {10.18653/v1/2025.findings-naacl.420},
  url       = {https://aclanthology.org/2025.findings-naacl.420/}
}

@article{yang2025moosechem2,
  title   = {{MOOSE-Chem2}: Exploring {LLM} Limits in Fine-Grained Scientific Hypothesis Discovery via Hierarchical Search},
  author  = {Yang, Zonglin and Liu, Wanhao and Gao, Ben and
             Liu, Yujie and Li, Wei and Xie, Tong and
             Bing, Lidong and Ouyang, Wanli and Cambria, Erik and
             Zhou, Dongzhan},
  journal = {arXiv preprint arXiv:2505.19209},
  year    = {2025},
  url     = {https://arxiv.org/abs/2505.19209}
}

\end{document}